\documentclass[sigconf]{acmart}
\AtBeginDocument{%
  \providecommand\BibTeX{{%
    \normalfont B\kern-0.5em{\scshape i\kern-0.25em b}\kern-0.8em\TeX}}}

\copyrightyear{2023}
\acmYear{2023}
\setcopyright{acmlicensed}\acmConference[ASIA CCS '23]{ACM ASIA
Conference on Computer and Communications Security}{July 10--14,
2023}{Melbourne, VIC, Australia}
\acmBooktitle{ACM ASIA Conference on Computer and Communications
Security (ASIA CCS '23), July 10--14, 2023, Melbourne, VIC, Australia}
\acmPrice{15.00}
\acmDOI{10.1145/3579856.3582819}
\acmISBN{979-8-4007-0098-9/23/07}

\usepackage{xspace}
\usepackage{algorithm}
\usepackage{algorithmic}
\usepackage{xspace}
\usepackage{amsmath}
\usepackage{enumitem}
\usepackage{caption}
\usepackage{subcaption}
\usepackage{multirow}
\usepackage{soul}
\newcommand{\methodName}{\textsc{RecUP-FL}\xspace}
\newcommand{\layerAtt}{Stru-NN}
\newcommand{\diffAtt}{Unkwn-NN}
\newcommand{\svmAtt}{SVM}
\newcommand{\rfAtt}{RF}

\DeclareMathOperator*{\argmin}{arg\,min}
\DeclareMathOperator*{\argmax}{arg\,max}

\newcommand{\yue}[1]{{\textcolor{cyan}{(Yue: #1)}}}

\newcommand{\BigO}[1]{\ensuremath{\operatorname{O}\bigl(#1\bigr)}}
\begin{document}

\title{\methodName: Reconciling Utility and Privacy in Federated Learning via User-configurable Privacy Defense}
\author{Yue Cui}
\affiliation{%
    \institution{University of Tennessee}
    \state{Knoxville, TN}
    \country{US}
    %\email{xxxx}
}
\email{ycui22@vols.utk.edu}

\author{Syed Irfan Ali Meerza}
\affiliation{%
    \institution{University of Tennessee}
    \state{Knoxville, TN}
    \country{US}
    %\email{xxxx}
}
\email{smeerza@vols.utk.edu}

\author{Zhuohang Li}
\affiliation{%
    \institution{University of Tennessee}
    \state{Knoxville, TN}
    \country{US}
    %\email{xxxx}
}
\email{zli96@vols.utk.edu}

%\author{Luyang Liu}
%\affiliation{%
%    \institution{Google Research}
%    \state{Mountain View}
%    \country{US}
    %\email{xxxx}
%}
%\email{luyangliu@google.com}

\author{Luyang Liu}
\affiliation{%
    \institution{Rutgers University}
    \state{New Brunswick, NJ}
    \country{US}
    %\email{xxxx}
}
\email{luyang@winlab.rutgers.edu}

\author{Jiaxin Zhang}
\affiliation{%
    \institution{Intuit AI Research}
    \state{Mountain View, CA}
    \country{US}
    %\email{xxxx}
}
\email{jiaxin_zhang@intuit.com}

\author{Jian Liu}
\affiliation{%
    \institution{University of Tennessee}
    \state{Knoxville, TN}
    \country{US}
    %\email{xxxx}
}
\email{jliu@utk.edu}

\renewcommand{\shortauthors}{Yue et al.}

\begin{abstract}
Federated learning (FL) provides a variety of privacy advantages by allowing clients to collaboratively train a model without sharing their private data. However, recent studies have shown that private information can still be leaked through shared gradients. To further minimize the risk of privacy leakage, existing defenses usually require clients to locally modify their gradients (e.g., differential privacy) prior to sharing with the server. While these approaches are effective in certain cases, they regard the entire data as a single entity to protect, which usually comes at a large cost in model utility.
In this paper, we seek to reconcile utility and privacy in FL by proposing a user-configurable privacy defense, 
\methodName, that can better focus on the user-specified sensitive attributes while obtaining significant improvements in utility over traditional defenses.
Moreover, we observe that existing inference attacks often rely on a machine learning model to extract the private information (e.g., attributes). We thus formulate such a privacy defense as an adversarial learning problem, where \methodName generates slight perturbations that can be added to the gradients before sharing to fool adversary models. To improve the transferability to un-queryable black-box adversary models, inspired by the idea of meta-learning, \methodName forms a model zoo containing a set of substitute models and iteratively alternates between simulations of the white-box and the black-box adversarial attack scenarios to generate perturbations.
Extensive experiments on four datasets under various adversarial settings (both attribute inference attack and data reconstruction attack) show that \methodName can meet user-specified privacy constraints over the sensitive attributes while significantly improving the model utility compared with state-of-the-art privacy defenses.

\end{abstract}

\vspace{-2mm}

%\ccsdesc[500]{Security and privacy~Privacy protections}

%\ccsdesc[500]{Computer systems organization~Embedded systems}

%\ccsdesc[500]{Security and privacy}
%\ccsdesc[500]{Computing methodologies }
%\ccsdesc[100]{Machine Learning}


 % \caption{Seattle Mariners at Spring Training, 2010.}
 % \Description{Enjoying the baseball game from the third-base
 % \label{fig:teaser}


\maketitle
\vspace{-4mm}
\section{Introduction}
\label{sec:intro}

% Benefiting from the recent surge in artificial intelligence technologies, we have witnessed a significant
% %unprecedented
% number of intelligent applications being developed in the past few years.
% As the new frontier of artificial intelligence,
Over the past few years,
deep learning models are being integrated into more and more mobile and edge/IoT applications to bring convenience to the users and help improve the user experience.
Successful applications can be found in almost every business sector, including but not limited to personal shopping recommendation~\cite{zheng2021personalized}, 
% mobile keyboard prediction~\cite{hard2018federated},
speech recognition~\cite{guliani2021training},
smart healthcare~\cite{apple_health}, and fraud prevention in mobile banking~\cite{bank_fraud}.
However, to power these intelligent applications, massive data need to be gathered from end users, which would inevitably cause privacy concerns.

Federated learning (FL)~\cite{mcmahan2017communication}, an emerging platform for distributed machine learning, has recently received considerable attention for its privacy benefits. In a typical FL system, a central \textit{server} coordinates multiple data providers (i.e., \textit{clients}) to collaboratively train a machine learning model. To protect privacy, clients do not directly share their private data during this learning process. Instead, the server and the clients exchange focused model updates (e.g., gradients) to achieve the learning objective. While offering practical privacy improvements over traditional centralized learning schemes, there is still no formal privacy guarantee in this vanilla form of FL~\cite{kairouz2021advances}. In fact, prior research has shown that different levels of private information may still be leaked through the shared model updates, ranging from membership information~\cite{melis2019exploiting}, sensitive attributes~\cite{lyu2021novel,feng2021attribute}, to even complete reconstruction of private training data samples~\cite{zhu2019deep,geiping2020inverting}.

In response to these threats, several privacy-preserving techniques have been proposed. For instance, secure multiparty computation (MPC)~\cite{danner2015fully,bonawitz2017practical} seeks to leverage cryptographic solutions to secure ``\textit{how it is computed}'' so that only the results of the computation are revealed to the intended parties.
%mohassel2017secureml
%For instance, secure multiparty computation (MPC)~\cite{danner2015fully,mohassel2017secureml,bonawitz2017practical} seeks to leverage cryptographic solutions to secure the computation process so that only the results of the computation are revealed to the intended parties. 
%However, such crypto-based solutions would induce heavy computational overheads. Moreover, 
%However, it has been shown that MPC alone is still not sufficient to resist certain inference attacks~\cite{melis2019exploiting,truex2019hybrid} and thus in practice, MPC is usually deployed in combination with other privacy-preserving techniques that can protect ``\textit{what is computed}'' (i.e., model updates shared by participating clients).
In addition, given the potential threats from other clients or malicious eavesdroppers on the client's communication channel, a formal privacy guarantee is needed on such client-level basis to protect ``\textit{what is computed}'' (i.e., model updates shared by participating clients).
For instance, to prevent gradient leakage without requiring trust in a centralized server, existing strategies require clients to apply a \textit{``local transformation''} to their gradients before sharing with the server, such as applying local differential privacy (local-DP)~\cite{geyer2017differentially,wei2021gradient}, gradient compression~\cite{zhao2019adversarial} and representation perturbation~\cite{sun2021soteria}, etc.

%For instance, differential privacy (DP) can introduce a level of uncertainty (e.g., adding random noises) into the shared updates~\cite{geyer2017differentially,wei2021gradient} to make the contribution of any individual client statistically indistinguishable.
%Other studies also showed that privacy can be preserved through gradient clipping~\cite{wei2021gradient}, gradient compression/sparsification~\cite{zhao2019adversarial} and representation perturbation~\cite{sun2021soteria}.

%such as differential privacy (DP). DP aims to limit and quantify the privacy disclosure about individuals by introducing a level of uncertainty (e.g., adding random noises) into the shared updates~\cite{geyer2017differentially,wei2021gradient}. DP can provide a rigorous mathematical guarantee so that the contribution of any individual client becomes statistically indistinguishable. Besides DP, there also exist other solutions attempting to protect clients' data privacy through perturbing the model updates, such as gradient sparsification~\cite{zhao2019adversarial} and representation perturbation~\cite{sun2021soteria}.

%\textbf{Limitations of Existing Efforts.}
\noindent\textbf{\underline{Limitations of Existing Efforts}.}
Despite offering remarkable privacy improvement, existing client-level approaches usually require adding a significant amount of noise (e.g., local-DP~\cite{geyer2017differentially,wei2021gradient}) or largely modifying the gradients~\cite{zhao2019adversarial,sun2021soteria}, which will inevitably depreciate the utility and usability of the resulting model.
Moreover, it is usually more challenging to maintain a reasonable utility-privacy trade-off with these client-level approaches. For instance, local-DP requires adding much higher noise than what is required by central-DP\footnote{Local-DP requires a lower bound of noise magnitude $\Omega(\sqrt{n}/\epsilon)$, while the central-DP only requires $\BigO{1/\epsilon}$, where $n$ is the number of participating clients and $\epsilon$ is the privacy loss~\cite{xiong2020comprehensive}.}.
On the other hand, existing solutions consider each client's private data as a single entity and attempt to protect all attributes, even including the ones that are helpful for the target learning task, and therefore they often come at a large cost in model utility. 
%To achieve better trade-offs between privacy and utility in FL, a possible direction is to consider appropriate relaxations on privacy/utility requirements~\cite{kairouz2021advances}.

%Despite offering remarkable privacy improvement, we observe that the existing privacy-preserving techniques in FL share the following limitations:
%(1) \textit{Current solutions can only achieve a suboptimal privacy-utility trade-off.} For instance, in practice, in order to completely hide the contribution of an individual, DP-based solutions often require adding a significant amount of noises, which will inevitably depreciate the utility and usability of the resulting model.
%(2) \textit{It is hard for current solutions to comply with existing privacy regulations.} In a practical FL scenario, the participating clients as data providers usually only agree on the data usage for the target training task (e.g., speech command recognition), while prohibiting other unintended sensitive information (e.g., voice biometric) to be used or leaked. Such right to control the usage of their contributed personal data is granted by recent data protection regulations, such as General Data Protection Regulation (GDPR)~\cite{GDPR}, California Consumer Privacy Act (CCPA)~\cite{CCPA}, and General Personal Data Protection Act (LGPD)~\cite{LGPD}, which mandate online services to take user consent before collecting, processing and sharing user's data.

%\textbf{Key Insights.}
\noindent\textbf{\underline{Key Insights}.}
To improve the utility-privacy trade-offs of privacy defenses in FL, we draw inspirations from the following key insights:
(1) Users may value different aspects of privacy differently, which may result in different privacy requirements~\cite{orekondy2017towards}. For example, people may have different comfort level sharing certain attributes, such as their political view, sexual orientation or religion.
%sharing images with partial nudity (i.e., subjects appear in undergarments) might be acceptable for some users, but less comfortable for other users. 
Thus, the actual privacy requirements can be relaxed by allowing user-specific privacy configurations.
(2) Recent data protection regulations, such as General Data Protection Regulation (GDPR)~\cite{GDPR}, and California Consumer Privacy Act (CCPA)~\cite{CCPA}, require giving data providers (e.g., clients in FL) their explicit consent to collect and process their sensitive personal data.
GDPR also makes a clear distinction between sensitive personal data and non-sensitive personal data, and sensitive data have more stringent requirements in terms of data collection and processing. 
While the non-sensitive data do not need to be treated with extra security, existing privacy defenses in FL regard all data equally sensitive and aim to sanitize the gradients to protect the whole data. Thus, we believe there is still much room for further reconciling utility and privacy preservation in FL.
(3) Most privacy leakages through gradients are caused by inference attacks, which usually rely on a machine learning model to learn the mapping between the exchanged model updates and private attributes.
On the other hand, learning models have been proven to be naturally vulnerable to adversarial examples~\cite{goodfellow2015explaining}, which can potentially be leveraged to mitigate such privacy leakage.

%In order to design a more practical privacy defense to address the aforementioned limitations, we draw inspirations from the following key insights:
%(1) Existing defense solutions often consider client's private data as a single entity and attempt to launch a general defense by assigning equal importance to all sensitive attributes, even including the ones that are helpful for the target training task, thereby degrading the utility of the model.
%(2) In contrast, in practice, different users may interpret the concept of privacy differently, which would result in different privacy preferences~\cite{orekondy2017towards}. For example, sharing images with partial nudity (i.e., subjects appear in undergarments) might be acceptable for some users, but less comfortable for other users with certain religions. Thus, allowing user-specific privacy configurations is highly desirable.
%(3) Most privacy leakages through gradients are caused by inference attacks, which usually rely on a machine learning model to infer private attributes from the exchanged model updates. On the other hand, learning models have been proven to be naturally vulnerable against adversarial examples~\cite{goodfellow2015explaining}, which can potentially be leveraged to mitigate such privacy leakage.

%\textbf{Our Solution.}
\noindent\textbf{\underline{Our Solution}.}
Based on these insights, in this paper, we propose \methodName, the first user-configurable local privacy defense framework that seeks to reconcile the utility and privacy in FL. 
%\luyang{Can we provide some example usecases of this kind of protection in intro?}
Unlike existing solutions that attempt to protect clients' entire training data, our objective is to focus on protecting \textit{a subset of} sensitive attributes specified by each user (i.e., client) according to their privacy preferences. 
%For example, in a FL speech recognition system, although many personal information can be revealed from the user's voice (e.g., speech content, geographical origin, emotions, age, and gender), users may be only sensitive to few of them.
By relaxing the privacy constraints on the non-sensitive attributes, \methodName can achieve a relaxed notion of privacy that better focuses on the identified sensitive attributes and at the same time obtain significant improvements in model utility over traditional defenses. 
For instance, voice data carry a set of sensitive information besides speech contents, such as identity, personality, geographical origin, emotions, gender and age, etc~\cite{feng2021attribute}. Users would value the privacy of these attributes differently given where and how their voice-controllable devices are used.
%Another example is much sensitive information such as biometrics (fingerprints, retina scan, voice signature, and facial geometry) are carried in clinical data, which may expose the identity of patients if left with no extra protection.
Similarly, images also contain different types of sensitive information, such as visited location, nationality, and fingerprint, etc. A wide user study conducted by Orekondy \textit{et al.}~\cite{orekondy2017towards} shows that most people think the leakage of fingerprint extremely violates their privacy while nationality is the least private information.
%When several hospitals cooperate in training cross-institutional models, the sensitive information may be considered as “Personally Identifiable Information” (PII)~\cite{alnemari2019protecting} and expose the identity of patients if left with no extra protection.
RecUP-FL provides a means for users to select any sensitive attributes that they would like to protect from all carried information before participating in the training, which can enhance privacy and raise their willingness to get involved in the training.

To achieve maximized privacy (i.e., reduce the attack success rate of attackers who leverage learning models to launch inference attacks as much as possible), \methodName is designed to be a local defense solution: at each communication round in FL, besides computing the model updates, each client also locally computes a perturbation based on the specified sensitive attributes and only shares the perturbed model updates to the server. In this way, the clients can ensure their targeted data privacy without trusting any other parties.
To protect specified sensitive attributes while maintaining a good level of utility, for each user-specified sensitive attribute, we formulate the defense as an optimization problem where the goal is to find the minimal perturbation that can prevent the adversary from making the correct prediction. Such formulation is equivalent to launching an adversarial attack against the adversary model that aims to classify the victim's sensitive attributes from the shared model updates.
However, computing such perturbations is not trivial, since the clients (1) possess no information about the configuration of the adversary model, including model parameters, model architecture, and even model type (e.g., Random Forest or Neural Network); and (2) do not have any query access to the adversary model, which makes existing query-based black-box adversarial attack methods~\cite{ilyas2018black} inapplicable.%,

To tackle the aforementioned challenges, \methodName adopts a two-stage method for calculating the defensive perturbation. In the first stage, \methodName forms a model zoo by loading a set of pre-trained substitute models (referred to as \textit{defender models}). For each selected sensitive attribute, the defender models mimics the behavior of the adversary by attempting to infer the attribute from the clients' model updates.
% The configuration of the defender models is carefully selected to cover a wide variety of model architectures to improve the defense's generalizability.
In the second stage, \methodName obtains the perturbations by launching an adversarial attack against the defender models. To enable the defense to be generalizable to unseen adversary models, \methodName exploits the meta gradient adversarial attack~\cite{yuan2021meta} to improve the transferability of the calculated adversarial perturbations.
% by iteratively alternating between simulations of the white-box and the black-box adversarial attack scenarios.
%\textcolor{red}{Specifically, inspired by the intuition of meta-learning \st{that learns knowledge from seen tasks and adopts quickly to unseen tasks}~\cite{nichol2018first}, for each iteration, \methodName first samples a subset of defender models from the model zoo to form the meta-train and meta-test tasks. In the meta-train steps, \methodName launches a white-box adversarial attack (i.e., FGSM~\cite{goodfellow2014explaining}) against an ensemble of defender models to obtain a temporary adversarial example, which is later used to compute the perturbation in the meta-test step by simulating a black-box attack scenario. Finally, the computed perturbations in the meta-test step of each iteration are accumulated to form the final perturbation.}
Note that \methodName targets ex-post empirical privacy instead of providing a formal differential privacy guarantee.
%\textcolor{blue}{Irfan: we can summarize this part in one or two sentences as we have a similar explanation in the design section as well}
We compare \methodName with four state-of-the-art baseline privacy defenses, including applying local differential privacy with both Gaussian noise~\cite{wei2021gradient} and Laplace noise~\cite{liu2020adaptive},
%gradient clipping~\cite{geyer2017differentially}, 
gradient sparsification~\cite{lin2017deep}, and Soteria~\cite{sun2021soteria}, under different settings with various types of threat models.
We consider two settings of threat models: (1) \textit{a third party eavesdropping on the communication channel} and (2) \textit{an honest but curious central server}. Both of them are able to infer the potential sensitive attribute information without affecting the training process.
The results show that the proposed \methodName can maximize the model utility and satisfy user-specified privacy constraints against various privacy attacks. 

\noindent\textbf{\underline{Contributions}.}
We summarize our main contributions as follows:

\begin{itemize}[leftmargin=*]
    \vspace{-1mm}
    \item To the best of our knowledge, \methodName is the first framework that seeks to reconcile utility and privacy via user-configurable local privacy defenses in FL .
    
    \item To improve utility and privacy trade-off, \methodName finds the minimal perturbation for protecting user-specified attributes by generating adversarial examples against a set of substitute defender models.
    
    \item In order to improve the generalizability of \methodName over unseen and un-queryable adversary models, we exploit the meta gradient adversarial attack method to iteratively improve the transferability of the defense by leveraging a collection of carefully-configured defender models.
    
    \item We evaluate the proposed \methodName on four datasets, including AudioMNIST, Adult Income, LFW and CelebA, under various adversary  settings. We show that \methodName is able to resist both attribute inference and data reconstruction attacks while achieving better utility-privacy trade-offs.
\end{itemize}

\vspace{-4mm}
\section{Related Work}
\vspace{-1mm}
\subsection{Privacy Leakage in Federated Learning}
\noindent \textbf{\underline{Attribute Inference Attack}.}
Attribute inference attack aims to infer certain input attributes of the client's private training data 
through analyzing shared gradient information.
This type of attack was first formulated in centralized learning against Hidden Markov Models (HMMs) and Support Vector Machine (SVM) classifiers~\cite{ateniese2015hacking} and then was extended to work on fully connected neural networks (FCNNs)~\cite{ganju2018property} to determine whether the training data has a certain set of properties.
In FL settings, Hitaj \textit{et al.}~\cite{hitaj2017deep} considered the adversary works as a client inside the privacy-preserving collaborative protocol and aims to infer class-representative information about a label that the adversary does not own. Further, Melis \textit{et al.}~\cite{melis2019exploiting} showed that an adversarial client can infer certain attributes of another client that are independent of its training task based on the exchanged model gradients (e.g., whether people in the training data wear glasses in a gender classification task). 
More recently, Lyu \textit{et al.}~\cite{lyu2021novel} considered a more practical scenario where clients share their epoch-averaged gradients instead of small batch-averaged gradients, and an honest but curious server will infer the sensitive attributes of local training data via a gradient-matching-based method.
Feng \textit{et al.}~\cite{feng2021attribute} proposed an attribute inference attack that can infer sensitive attributes (e.g., the client's gender information) from shared gradients while training a speech emotion recognition classifier via shadow training.

\noindent \textbf{\underline{Data Reconstruction Attack}.}
%The data reconstruction attack is the attackers take the updated gradients shared by the clients as input, and try to reconstruct the private training information.
Prior studies showed the possibility of recovering class-level~\cite{hitaj2017deep}  or even client-level~\cite{wang2019beyond} data representatives through generative models.
More recent studies~\cite{zhu2019deep,zhao2020idlg,geiping2020inverting,yin2021see} showed that an adversary could even fully restore the training data from its shared gradient information.
Specifically, Zhu \textit{et al.} recently~\cite{zhu2019deep} proposed to solve this gradient inversion problem by solving for the optimal pair of input and label that best matches the exchanged gradients. 
As a follow-up study, Zhao \textit{et al.}~\cite{zhao2020idlg} provided an analytical computation method to precisely infer the label information by performing binary classification to the direction of the last layer's gradient. It can effectively involve label information in the reconstruction process and thus improve the attack performance. However, they can only work on shallow network architectures with low-resolution images.
To launch such attacks in more realistic scenarios, Geiping \textit{et al.}~\cite{geiping2020inverting} proposed to use a magnitude-invariant design along with various optimization strategies to restore ImageNet-level high-resolution data in large batch size from deeper networks (e.g, ResNet~\cite{he2016deep}).
Yin \textit{et al.}~\cite{yin2021see} also achieved image batch reconstruction by utilizing batch normalization statistics and image fidelity regularization. 

\vspace{-4mm}
\subsection{Privacy Defenses in Federated Learning}
%FL can be considered as a \textit{computation} on a distributed client dataset. According to the privacy aspects of the computation, existing privacy defenses in FL can be broadly categorized into gradient-degradation-based (\textit{what is computed}) and crypto-based (\textit{how it is computed}) defenses.

%\textbf{Crypto-based Defenses.}
\noindent \textbf{\underline{Crypto-based Defenses}.}
One line of defense strategy is to protect the aggregation of model updates through secure multi-party computation (MPC)~\cite{danner2015fully,mohassel2017secureml,bonawitz2017practical}, where a set of parties jointly compute a common function of interest without revealing their private inputs to other parties. For instance, Danner \textit{et al.}~\cite{danner2015fully} proposed a secure sum protocol using a tree topology and homomorphic encryption.
SecureML~\cite{mohassel2017secureml} adopt a two-server model for preserving privacy, in which clients process, encrypt, and/or secret-share their data among two non-colluding servers to train a global model.
Additionally, Bonawitz \textit{et al.}~\cite{bonawitz2017practical} require the aggregation of model updates in FL to be logically performed by the virtual, incorruptible third party so that the server can only receive the aggregated model update. However, these crypto-based methods would inevitably cause high computational overhead, and recent inference attacks~\cite{melis2019exploiting} showed that the adversary can still reveal private information even though they can only access the aggregated model update. Therefore, to ensure rigorous privacy guarantees in FL, secure computation techniques is usually deployed in parallel with the techniques for privacy-preserving disclosure, such as gradient-degradation-based defenses~\cite{kairouz2021advances} to be mentioned next.

%\textbf{Gradient-degradation-based Defenses.}
\noindent \textbf{\underline{Gradient-degradation-based Defenses}.}
To prevent privacy leakage via shared gradient information in FL, a very straightforward way is to intentionally ``degrade'' the fidelity of gradients on the client prior to sharing them with the server.
As a standard and common method, differential privacy (DP) can be either applied at the client side (i.e., local DP) or the server side (i.e., central DP) to perturb the client's shared gradients and the aggregated gradients~\cite{geyer2017differentially,wei2021gradient}, respectively, and thereby mitigate privacy risks.
Compared with central DP, local DP usually provides a better notion of privacy since it does not require trust in a centralized server. However, local DP requires injecting random noises to the gradients at a large number of clients, making this local approach often come at a large cost in utility.
Zhao \textit{et al.}~\cite{zhao2019adversarial} theoretically and empirically proved that DP makes data private by adding a significantly large amount of noise, but it simultaneously filters out much useful information. 
In addition to DP, Zhu \textit{et al.}~\cite{zhu2019deep} demonstrated that performing gradient sparsification (i.e., gradients with small magnitudes are pruned to zero) can also help prevent privacy leakage from the gradient.
%Wei \textit{et al.}~\cite{wei2021gradient} showed that gradient clipping that constrains the norm of the gradients to a given bound could also be applied to defend against various attacks by restricting the influence of each model update. 
A more recent work, Soteria~\cite{sun2021soteria}, proposed to compute the gradients based on perturbed data representations to maintain a good level of model utility while achieving a certified robustness guarantee to FL.
While these gradient-degradation-based methods can mitigate privacy risks in certain cases, they can only achieve a sub-optimal utility-privacy trade-off because they treat the entire training data (including non-sensitive information) as a single entity, thereby redundantly degrading the gradients' fidelity.
Different from the above methods, \methodName only protects the sensitive attributes identified by users instead of the whole data, thus obtaining an improved utility-privacy trade-off.
The general framework of our approach is under the umbrella of context-aware privacy defenses (e.g.,~\cite{huang2017context,huang2018generative,raval2019olympus}), which can leverage the context (e.g., dataset statistics, dataset's utility) to achieve better utility-privacy trade-offs. 
%In contradict to context-free privacy defenses such as the above-mentioned DP, context-aware defenses leverage the dataset statistics to achieve an improved utility-privacy trade-off. 
%Huang~\textit{et al.}~\cite{huang2017context,huang2018generative} introduced a novel context-aware privacy framework that leverages generative adversarial networks (GANs) to learn the privatization schemes from the dataset itself.
%Raval~\textit{et al.}~\cite{raval2019olympus} proposed OLYMPUS to limit the privacy leakage. It modeled the privacy and utility requirements as adversarial networks and obfuscated private information while minimizing the influence on the functionality of data.
However, to the best of our knowledge, this is the first work that leverages the knowledge of user-specific sensitive attributes to improve utility-privacy trade-offs in FL.

\vspace{-2mm}
\section{Problem Formulation \& \methodName~Design Objectives}

%In this section, we formulate the privacy leakage problem by reviewing the background related to the process of FL in section~\ref{FL}, defining two of the most severe kinds of attacks that exist in FL formally in section~\ref{threat models}, discussing the limitations of existing defenses in section~\ref{privacy defense}. To address these issues, we propose \methodName and describe its design objectives in section~\ref{design objectives}.  

\subsection{Problem Formulation}~\label{subsec:problem}
\noindent\textbf{\underline{Federated Learning}.}
Without loss of generality, we assume there are $K$ participating clients $\mathbf{C} = \{C_1,..., C_K\}$ and one central server in our FL setting. The clients $\mathbf{C}$ will collaboratively train a global model $\mathcal{G}$ under the organization of the central server.
The client $C_i$ holds its local data $D_i$ which is composed by $N$ individual data records $(\mathbf{X}_i, \mathbf{Y}_i) = \{(x_{i,1}, y_{i,1}), ..., (x_{i,n},y_{i,n})\}, n \in [1,N]$, where $x_{i,n}$ denotes the $n$-th data sample in the $i$-th client, and $y_{i,n}$ denotes its corresponding label of the training FL task. 

%The clients and the server work jointly to obtain the optimal global model $M$ with weights $w^*$ to solve the optimization problem $w^*=argmin_{w} F(D,w)$, where $F(\cdot)$ denotes the loss on the global model.
% The central server will allocate the global model to clients, then each client will finetune the received model on its local dataset and send back the model update $\nabla w_{t,i}, i \in [1,K]$.
% Finally, the central server will collect and aggregate all the received model updates to update the global model. This process will be repeated until the global model is converged.
% To be specific, at the beginning of the FL process, a global model $M$ is first randomly initialized with weights $w_0$. After initialization, the central server repeatedly interacts with clients at each communication round until the model converges. A communication round at time $t \in [1, ..., T]$ contains the following steps:
At the beginning of the FL process, the central server first initializes the global model $\mathcal{G}$ with random initial weights $w_0$. After initialization, the central server repeatedly interacts with clients for $T$ communication rounds until the global model converges. Each communication round $t \in [1,T]$ contains the following steps:
\vspace{-2mm}
\begin{itemize}[leftmargin=*]
    \item \textbf{Step 1: Synchronization.} The central model sends the current model $\mathcal{G}$ with weights $w_{t}$ to all $K$ clients.
    \item \textbf{Step 2: Local Training.} Each client $C_i$ performs one or more training steps on the received model $\mathcal{G}$ using its local data $D_i$. After training, each client sends its model update (i.e., gradients) $\nabla w_{t,i}$ back to the central server.
    \item \textbf{Step 3: Aggregation.}
    The central server averages all participating clients' model updates to update the global model~\cite{mcmahan2017communication}: $w_{t+1} = w_{t} - \alpha \cdot \sum_{i=1}^{K} \frac{\nabla w_{t,i}}{K}$ via gradient descent, where $\alpha$ is the learning rate. 
    % The central server computes a global model update by an aggregation rule based on the clients' individual updates $\nabla w_{t,i}, i \in K$.
    % Unless mentioned otherwise, we adopt the most commonly used aggregator FedAvg~\cite{mcmahan2017communication}, which aggregates the clients' updates according to: $w_{t+1} = w_{t} + \alpha \sum_{i=1}^{K} \frac{1}{K} \nabla w_{t,i}$ where $\alpha$ is the learning rate of global model.
    
\end{itemize}

% = \frac{\partial \mathcal{L}(D_i, w_t)}{\partial w_t}，, where $\mathcal{L}(\cdot)$ is the loss function of the model $\mathcal{G}$

Unlike the centralized training scheme that requires clients to send their local data $D_i$ to the central server, FL only requires the model updates $\nabla w_{t,i}$ to be shared with the central server, and thereby the privacy concerns can be mitigated. 

\noindent\textbf{\underline{Threat Models}.}
Despite the fact that training data can be kept locally in FL, the shared model updates
%, which are computed based on the local data,
still carry much sensitive information about the local data, which can be leveraged by an adversary to gain knowledge of the client's private data.
% Although an adversary cannot access the private data itself, it is able to gather and analyze the model updates to be sent to the central server in such a way as to gain the desired knowledge of the client's private data.
In order to evaluate our defense under the worst-case scenario,
we assume a very powerful adversary with the ability to access each client's update. In practice, the adversary can be either (1) \textit{a third party} outside the training process eavesdropping on the communication channel~\cite{zari2021efficient}, gathering the model updates, and launching attacks, or (2) \textit{an honest but curious central server}, who executes the regular training procedure but also attempts to infer the client's private information from the received model updates~\cite{lyu2021novel}.
%\textcolor{red}{The adversary can be xxxx in reality.}
%who performs the aggregation rules and organizes the training honestly, but observes the model updates from some victim clients secretly and performs privacy leakage attacks to gain the knowledge of the client's training data at the same time.
The goal of the adversary is to reveal as much sensitive information about the client's private data as possible.

\begin{figure*}[t]
    \centering
    \includegraphics[width=0.9\linewidth]{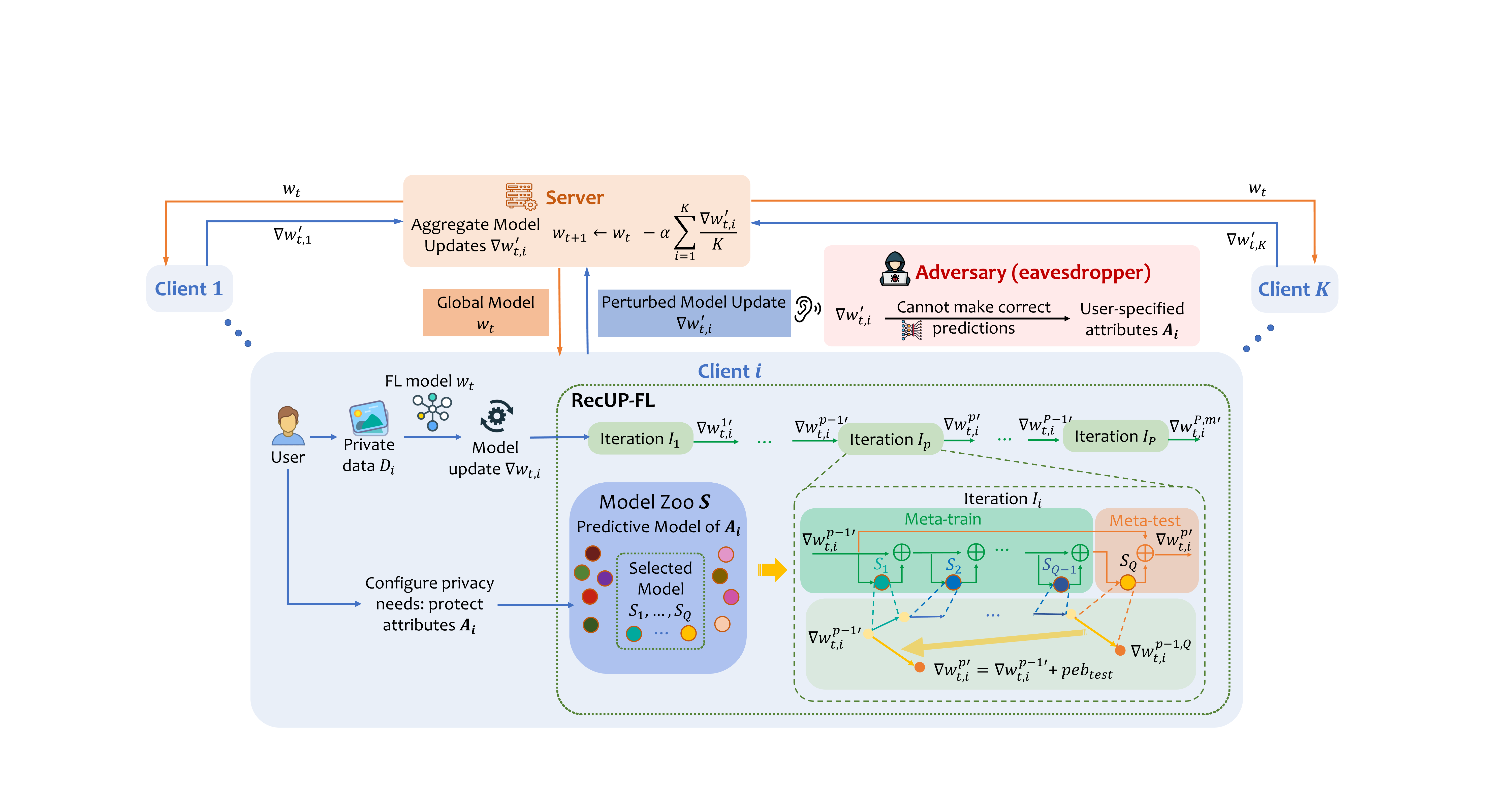}
    \vspace{-4mm}
    \caption{Overview of~\methodName.}
    \label{fig:workflow}
    
    \vspace{-4mm}
\end{figure*}

To investigate the worst-case scenario, we consider two critical privacy leakage attacks in FL: attribute inference attack and data reconstruction attack. In the attribute inference attack, the adversary aims to infer the sensitive attributes of the client's private training data from the shared model updates. As stated in prior studies~\cite{ganju2018property,shokri2017membership}, it usually builds an adversary model $\mathcal{I}$, intercepts the model updates $\nabla w_{t,i}$, and infers the sensitive attribute value $a_i$ utilizing the pre-trained adversary model (i.e., $a_i^{'}=\mathcal{I}(\nabla w_{t,i})$). In the data reconstruction attack, the adversary aims to completely recover the client's private training data $X_i$ from the shared model updates by solving a gradient-matching optimization problem~\cite{zhu2019deep, zhao2020idlg, geiping2020inverting} as $X_i^{'}=\mathcal{R}(\nabla w_{t,i}))$ where $\mathcal{R}$ denotes the reconstructor.

\noindent\textbf{\underline{Privacy Defense}.}
A common way to defeat inference attack in FL is to apply a defensive local transformation function $\varphi(\cdot)$ on the model update before sharing as follows:
\begin{equation}
\setlength{\abovedisplayskip}{3pt}
\setlength{\belowdisplayskip}{3pt}
    \nabla w_{t,i}^{\prime} = \varphi(\nabla w_{t,i}),
\end{equation}
where $\nabla w_{t,i}^{\prime}$ is the perturbed model update after applying the transformation.
We consider the following four state-of-the-art defensive transformation functions:

%\noindent (1) \textbf{DP (Gaussian)}~\cite{wei2021gradient}: One of the most common solutions to defend against privacy leakage attacks in FL is DP (Gaussian), which restricts the model updates within the given clipping bound and injects Gaussian noise on the model updates before sharing.
% Gaussian noise is a popular kind of noise to be added as in~\cite{zhu2019deep}.
%Given the mean $\mu$, standard deviation $\sigma$ and clipping bound $B$, the transformation function of DP (Gaussian) is $\varphi_g(\nabla w_{t,i}, \mu, \sigma, B) = \frac{\nabla w_{t,i} + \mathcal{N}(\mu, \sigma^{2})}{\max (1, \|\nabla w_{t,i} + \mathcal{N}(\mu, \sigma^{2})\|_2/B)}$, where $\mathcal{N}$ is a normal distribution.

\noindent (1) \textbf{DP (Gaussian)}~\cite{wei2021gradient}: One of the most common solutions to defend against privacy leakage attacks in FL is DP (Gaussian). It restricts the model updates within the given clipping bound $B$ by $ Clip(\nabla w_{t,i}, B) = \frac{\nabla w_{t,i}}{\max (1, \|\nabla w_{t,i}\|_2/B)}$. Then it injects Gaussian noise on the clipped model updates before sharing by $\varphi_g(\nabla w_{t,i}, \mu, \sigma, B) = Clip(\nabla w_{t,i}, B) + \mathcal{N}(\mu, \sigma^{2})$, where $\mathcal{N}(\cdot)$ is a normal distribution, $\mu$ and $\sigma$ are the mean and standard deviation of the noise.
% Gaussian noise is a popular kind of noise to be added as in~\cite{zhu2019deep}.
%Given the mean $\mu$, standard deviation $\sigma$ and clipping bound $B$, the transformation function of DP (Gaussian) is $\varphi_g(\nabla w_{t,i}, \mu, \sigma, B) = \frac{\nabla w_{t,i} + \mathcal{N}(\mu, \sigma^{2})}{\max (1, \|\nabla w_{t,i} + \mathcal{N}(\mu, \sigma^{2})\|_2/B)}$, where $\mathcal{N}$ is a normal distribution.

%\noindent (2) \textbf{Gradient Clipping}~\cite{geyer2017differentially}: As suggested in prior research~\cite{geyer2017differentially}, constraining the impact on individual update is an effective way in DP studies to defend against various types of attacks. Typically, the transformation of gradient clipping can be written as $\varphi_c(\nabla w_{t,i}, B) = \frac{\nabla w_{t,i}}{\max (1, \|\nabla w_{t,i}\|_2/B)}$, where $B$ is the given clipping bound. By doing this, the norm of the model updates can be restricted within the clipping bound $B$. 

%\noindent (2) \textbf{DP (Laplace)}~\cite{liu2020adaptive}: In addition to DP (Gaussian), DP (Laplace) can also be used to achieve differential privacy. Similarly, given the location $\mu$, the scale $b$ and clipping bound $B$, the transformation function of adding Laplace noise is $\varphi_l(\nabla w_{t,i}, \mu, b, B) = \frac{\nabla w_{t,i} + Lap(\mu, b)}{\max (1, \|\nabla w_{t,i} + Lap(\mu, b)\|_2/B)}$, where $Lap(\cdot)$ denotes the Laplace distribution.

\noindent (2) \textbf{DP (Laplace)}~\cite{liu2020adaptive}:
%In addition to DP (Gaussian), DP (Laplace) can also be used to achieve differential privacy.
Similar to DP (Gaussian), given the location $\mu$, the scale $b$ and clipping bound $B$, the transformation function of adding Laplace noise is $\varphi_l(\nabla w_{t,i}, \mu, b, B) = Clip(\nabla w_{t,i}, B) + Lap(\mu, b)$, where $Lap(\cdot)$ denotes the Laplace distribution.

%$\varphi_l(\nabla w_{t,i}, \mu, b) = \nabla w_{t,i} + Lap(\mu, b)$

\noindent (3) \textbf{Gradient Sparsification}~\cite{lin2017deep}: Gradient sparsification is originally proposed to reduce the communication cost in FL and is later proved to be also effective in defending against certain gradient leakage attacks~\cite{zhu2019deep}. Given a sparsification rate $p \in (0,1)$, a binary mask $M$ is first calculated by $\mathcal{M} \leftarrow \|\nabla w_{t,i} \|> p$  of $\|\nabla w_{t,i} \|$,
then the mask is applied to the original model update according to $\varphi_s(\nabla w_{t,i}, p) = \mathcal{M} \odot \nabla w_{t,i}$, where $\odot$ denotes the point-wise multiplication operation.
% then the transformation function $\varphi_s(\nabla w_{t,i}, p) = \mathcal{M} \odot \nabla w_{t,i}$ can be considered as applying a point-wise mask $\mathcal{M}$ on the original gradient in order to set the smallest $p$ positions of the gradient to zero.

\noindent (4) \textbf{Soteria}~\cite{sun2021soteria}: A recent solution to defend against data reconstruction attacks in FL. Soteria perturbs the representation of the input data that learned from a fully-connected layer $L$ (called the defend layer) to maximize the reconstruction error. Suppose the global model of FL consists of a feature extractor before the defend layer and the classifier denoted as $f_r$. $f_r$ first learns to map the target input data sample $x_{i, n} \in \mathbb{R}^d$ to a $l$-dimension representation $r \in \mathbb{R}^l$. Then, the classifier maps the learned representation $r$ to the classes of the training task. Specifically, given a pruning rate $p \in (0,1)$, the client first evaluates the impact $\iota_i$ of each element $r_i \in r$ by calculating $\iota_i = \|r_i(\nabla_{x_{i, n}} f_r (r_i))^{-1} \|_2$. Then, the client prunes the $p \times l$ elements in the defender layer to the largest value in ${\iota_i, i \in  [0,l-1]}$ to get a perturbed representation $r'$ of the input dataset $D_i$. Finally, the client computes and shares the update computed on the perturbed representation $r^{\prime}$. Therefore, the defensive transformation function of Soteria can be considered as applying a mask $\mathcal{M}$ only to the update of the defend layer, which can be written as $\varphi_{sot}(\nabla w_{t,i}, p) = \mathcal{M} \odot \nabla w_{t,i}$.

While the above-mentioned client-level defenses can prevent privacy leakage in certain cases, the applied local transformation $\varphi(\cdot)$ on the model update will inevitably degrade its fidelity and thereby greatly impacts the utility of the resulting model. Additionally, we can observe from their transformation functions that all of them treat the whole model update as an entire entity and regard all attributes as equally important. However, sensitive attributes usually have unequal emphases at different layers/positions of the gradients~\cite{mo2021quantifying}, and more importantly, users may value different aspects of privacy differently~\cite{orekondy2017towards}. 
Therefore, to meet actual privacy constraints, such a general treatment, which provides either redundant protection on non-sensitive gradients or insufficient protection on sensitive gradients, can hardly achieve an optimal utility-privacy trade-off.

%The above-mentioned defenses apply different kinds of transformation functions $\varphi(\cdot)$ on the update before sharing, we find that they share the following limitations:

%(1) \textit{Existing efforts can only achieve a sub-optimal privacy-utility trade-off.} We can observe from their transformation functions that all of them treat the whole gradient as an entire entity and apply a general transformation to all sensitive attributes. However, sensitive attributes have unequal emphases at different positions of the gradient. Therefore, to achieve sufficient privacy protection, such a general treatment provides redundant protection for those unimportant positions and insufficient protection for the more important positions.

%(2) \textit{Existing solutions are hard to comply with recently updated privacy regulations.} The latest privacy regulations, such as GDPR~\cite{GDPR}, CCPA~\cite{CCPA} and LGPD~\cite{LGPD}, require more flexible control over the data to be provided to clients. In the FL scenario, in addition to the agreement on the data usage for the FL training task, the training organizer (i.e., the central server) should also make sure other sensitive attributes will not be leaked. However, existing studies are not able to provide such control since they only focus on the training task. 

\vspace{-4mm}
\subsection{Design Objectives}
\label{design objectives}

To address the limitations of existing efforts, we propose \methodName, a user-configurable local privacy defense framework that seeks to reconcile the utility and privacy in FL. 
% In practice, in addition to the label of training task $y_{i,n}$, each data record $x_{i,n}$ also has a set of sensitive attributes. The clients can specific a subset of all attributes $\mathbf{A_i}=\{a_{i,1}, ..., a_{i,m}\}$ to protect.
Unlike existing approaches, \methodName can achieve a relaxed notion of privacy by providing effective protection only on the sensitive attributes specified by each user (i.e., client) according to their preference.
This way the defense can be more flexible and can better adjust to different users’ privacy requirements in practice while obtaining significant improvements in model utility.
%This way the defense can meet the user's actual privacy requirements and meanwhile obtain significant improvements in model utility.
Specifically, 
%In addition to the label of training task $y_{i,n}$, 
\methodName allows each client to specify a set of attributes $\mathbf{A}_i=\{a_{i,1}, ..., a_{i,M}\}$ to protect its local training data before the FL training.
%record $x_{i,n}$.
%Only providing protection on the specific attributes subset, \methodName can obtain significant improvements in model performance.
To protect the user-specified attributes, \methodName will leverage the idea of adversarial learning to find the optimal (minimal) perturbation $p_{t,i}$ to be added to the original model update $\nabla w_{t,i}$ for the client $C_i$ at the $t$-th communication round before sharing. This can be formally described as $\varphi_\textup{\methodName}(\nabla w_{t,i}) = \nabla w_{t,i} + p_{t,i}$.
The objectives of \methodName are twofold: 

\begin{itemize}[leftmargin=*]
    \item \textbf{Objective I (Privacy)}: The perturbed model update should be able to prevent the adversary from inferring the user-specified sensitive attributes.
    \item \textbf{Objective II (Utility)}: \methodName should find the minimal perturbation to maintain a good level of utility of the global model.
\end{itemize}

%To formulate these two objectives, we have the following two objective functions.

To achieve \textbf{Objective I},
we require the attributes inferred from the perturbed updates to be as different from the genuine attributes as possible:
% we require the predicted user-configured attributes of each client should be as dissimilar to the real values as possible.
% The corresponding objective function can be written as:
\vspace{-2mm}
\begin{equation}
\setlength{\abovedisplayskip}{3pt}
\setlength{\belowdisplayskip}{3pt}
    \argmax_{p_{t,i}} d(\mathcal{F}(p_{t,i}+\nabla w_{t,i}), \mathbf{A}_i),
\end{equation}
where $d(\cdot)$ measures the distance between the inferred and genuine attributes and $\mathcal{F}(\cdot)$
denotes the adversarial prediction function, which takes model update as input and outputs the predicted sensitive attributes.
That is, instead of attempting to degrade the reconstructed data quality, we only aim to prevent the updates from leaking information about the user-specified sensitive attributes.
%Specifically, for attribute inference attacks, $\mathcal{F}$ is equivalent to the adversary model $\mathcal{I}$.
%For data reconstruction attack, $\mathcal{F}$ is equivalent to the adversary's data reconstructor $\mathcal{R}$.
%\textcolor{red}{refer to deleted formula}

% To be specific, if the adversary launches the attribute inference attack, the prediction function $\mathcal{F}(\cdot)$ is the pre-trained gradient classifiers $\mathcal{I}$.
%For data reconstruction attack,
%$\mathcal{F}(p_{t,i}+\nabla w_{t,i}) = \mathcal{P}(\mathcal{R}(p_{t,i}+\nabla w_{t,i}))$, where $\mathcal{R}$ is the adversary's data reconstructor and $\mathcal{P} = \{\mathcal{P}_1, ..., \mathcal{P}_M\}$ is a set of auxiliary attribute classifiers that try to predict the sensitive attributes by observing the recovered data samples.

% If the adversary launches the data reconstruction attack, the $\mathcal{F}(\cdot)$ will be a combination of a reconstruction function $\mathcal{R}(\cdot)$ and a set of classifiers $\mathcal{P} = \{\mathcal{P}_1, ..., \mathcal{P}_m\}$ as $\mathcal{F}(\cdot) = \mathcal{P}(\mathcal{R}(\cdot))$.

% It means instead of requiring the reconstructed data to become dummy data, we only require the reconstructed data will not leak the user-configured attributes information. That is, the attribute classifiers $\mathcal{P}$ cannot make correct predictions on the reconstructed data. 

To achieve \textbf{Objective II}, we seek to minimize the global model's training loss: 

\vspace{-4mm}
\begin{equation}
\setlength{\abovedisplayskip}{3pt}
\setlength{\belowdisplayskip}{3pt}
    \argmin_{p_{t,i}} \mathcal{L}(w_t+\alpha \sum _{i=1}^{N} (p_{t,i}+\nabla w_{t,i}), \ D_{test}),
\end{equation}
where $\mathcal{L}(\cdot)$ is the training loss function (e.g., cross-entropy) and $D_{test}$ is the evaluation set.

\vspace{-3mm}
\section{Design of~\methodName}

\subsection{System Overview}
As shown in Figure~\ref{fig:workflow}, the proposed \methodName can be implemented as an add-on defense module on the client side to prevent privacy leakage. To improve model utility while still meeting user's privacy requirements, \methodName provides users with more control over their privacy and allows each client to configure their privacy needs before the FL training by identifying a set of sensitive attributes (i.e., $\mathbf{A_i}$ for the $i$-th client).
In practical applications, the sensitivity and configuration of attributes only rely on users' personal preferences instead of any external regulations. For example, for most people, voice biometrics in general are often considered more private than emotion in speech data. They may only identify voice biometrics as a sensitive attribute to protect.
This way we can achieve a relaxed notion of privacy that only needs to meet users' actual privacy needs.
Specifically, in FL, at the communication round $t$, the $i$-th client needs to calculate the model update $\nabla w_{t,i}$ first through learning its private data with the FL model $w_t$. To protect the identified sensitive attributes while maintaining a good level of utility, we formulate the defense as an adversarial machine learning attack problem that seeks to find a minimal perturbation to be added to the model update before sharing to mislead the adversary model so that the adversary cannot reveal any private information from the perturbed update, i.e., $\nabla w_{t,i}^{\prime}$. Note that \methodName does not require the server to execute any additional tasks besides aggregation. We use FedAvg~\cite{mcmahan2017communication} in this work for simplicity, but in practice, \methodName can also work with other secure aggregation rules, such as Krum~\cite{blanchard2017machine}, and Median~\cite{yin2018byzantine}, etc.

Although the client-level privacy defense can be formulated as launching an adversarial attack against the privacy-leakage adversary model, solving such an optimization problem to generate gradient perturbations is not trivial. Unlike existing adversarial attacks (mostly in white-box~\cite{goodfellow2015explaining, carlini2017towards} or black-box~\cite{ilyas2018black} settings), for the privacy defense purposes in FL, we need to consider a very restricted no-box setting as clients do not possess any knowledge about the configuration of the adversary model (e.g., model architecture and parameters) and, more importantly, they are not able to query the adversary model to counterfeit its functionality. Additionally, existing studies~\cite{melis2019exploiting, jia2018attriguard} showed the feasibility of using non-neural-network approaches (e.g., Random Forest and Support Vector Machines) to infer sensitive attributes from model parameters. To make our gradient perturbation applicable to arbitrary adversary models regardless of their model type, architecture, and parameters, we need to improve the cross-model transferability of our generated defensive gradient perturbations. %weinsberg2012blurme

Meta-learning~\cite{nichol2018first} is proposed for solving unseen tasks by \textit{learning to learn}. A meta-learning model first learns knowledge and seeks the inner connections from multiple training tasks (i.e., meta-train). Then later the model is adapted to the unseen task by fine-tuning with only few training samples (i.e., meta-test). A detailed introduction to meta-learning can be founded in Appendix~\ref{sec:meta-learning}.
Inspired by the meta-learning technique, we propose to use a two-step iterative method to generate perturbations for unseen and unquerable adversary models.
As shown in Figure~\ref{fig:workflow}, at each iteration, we have: (1) \textit{Meta-learn}: using a set of known defender models (i.e., $\mathbf{S}$), as substitutes of the adversary model, to sequentially generate a \textit{universal} adversarial perturbation (equivalent to launching white-box attacks); and (2) \textit{Meta-test}: leveraging the prior knowledge in its learned universal perturbation to fine-tune itself to a new unseen defender model (equivalent to launching black-box attacks). 
By iteratively conducting white-box and black-box attacks, \methodName can narrow the gap between the gradient directions in white-box and black-box attacks and gradually learn to adjust the perturbation from known defender models to the unknown adversary model. Further, we repeat the above two-step iterative method several times. The different compositions of substitute models each time make the generated perturbation will not bias to any specific model.

\vspace{-4mm}
\subsection{Methodology}
\label{sec:methodology}
Suppose \methodName generates defensive perturbations to protect the model update $\nabla w_{t,i}$ calculated by the $i$-th client at the $t$-th communication round, and the client specifies its sensitive attribute set $\mathbf{A}_i = \{a_{i,1}, ..., a_{i,M}\}$. For simplicity, we first use the single attribute protection as an example to introduce our method and then expand it multi-attribute protection. The complete process of generating defensive perturbations can be found in Appendix Algorithm~\ref{algo}. A theoretical analysis can be found in Appendix Section~\ref{sec:theoretical}.

%Suppose we are trying to protect the model update $\nabla w_{t,i}$ from the $i$-th client at the $t$-th communication round. And the $i$-th client specify an attribute subset $\mathbf{A}_i = \{a_{i,1}, ..., a_{i,M}\}$ to protect.
%, our adversarial sample should be able to mislead a set of adversary models that predict $a_{i,1}, ..., a_{i,m}$. To simplify, we use the single attribute protection as an example to introduce our method in section~\ref{single} and then expand it to multiple attributes protection in section~\ref{multiple}. The complete process can be found in Algorithm~\ref{algo}.
\noindent\textbf{\underline{Single Attribute Protection}.}
In this case, we consider $a_{i,m} \in \mathbf{A}_i$ as the targeted attribute to provide protection. Before the FL training, a model zoo, $\mathcal{S}$, consisting of multiple diverse pre-trained substitute models (referred to as defender models) needs to be created. These models all mimic the behavior of the adversary model, that is, predicting the $a_{i,m}$ through model updates. In practice, we can create an ensemble of models for every possible sensitive attribute prior to the FL training and pre-load the corresponding ensemble models to the clients according to their attribute specifications.
To improve the transferability of the defensive perturbations to the unseen and un-queryable adversary model, we randomly sample $\{S_1, ..., S_Q\}$ from $\mathcal{S}$ and perform the following \textit{meta-train} and \textit{meta-test} to compute defensive gradient perturbations. 

\vspace{-2mm}
\begin{itemize}[leftmargin=*]
\item \textbf{Meta-train.} Meta-train utilizes the first $Q$-$1$ models from the selected $Q$ models to simulate white-box adversarial attacks to generate defensive perturbations. Considering clients usually only have limited computational resources, we adopt the Fast Gradient Sign Method (FGSM)~\cite{goodfellow2015explaining} as our white-box attack method due to its fast execution and ``free'' adversarial training features compared to other computationally expensive attacks such as projected gradient decent (PGD)-based attacks~\cite{carlini2017towards}.%moosavi2016deepfool,madry2017towards
%Since our method can be integrated with any gradient-based attack method, we adopt the fast gradient sign method (FGSM)~\cite{goodfellow2014explaining} as our white-box attack method because of its low computation cost.
Specifically, FGSM directly utilizes the gradient information of the targeted model by modifying the benign input to the opposite direction of the correct prediction. 
The perturbation generation in FGSM can be described as:
\begin{equation}
\setlength{\abovedisplayskip}{3pt}
\setlength{\belowdisplayskip}{3pt}
    \nabla w_{t,i}^{q} = \nabla w_{t,i} + \frac{\epsilon}{Q} \cdot sign(\nabla g_q(S_q(\nabla w_{t,i}),a_{i,m})),
\end{equation}
where $\nabla w_{t,i}^{q}$ denotes the perturbed model update that aims to fool the defender model $S_q$, $\epsilon$ denotes the perturbation budget, and $g_q(\cdot)$ denotes the loss function of the model $S_q$. 
%FGSM only takes one step to mislead the adversary model, which is light enough to apply to any personal device.
To generate a universal perturbation that can be applied to the first $Q$-$1$ defender models, we employ the iterative FGSM as:
\begin{equation}
\setlength{\abovedisplayskip}{3pt}
\setlength{\belowdisplayskip}{3pt}
    \nabla w_{t,i}^{q} = \nabla w_{t,i}^{q-1} + \frac{\epsilon}{Q} \cdot sign(\nabla g_q(S_q(\nabla w_{t,i}^{q-1}), a_{i,m})), q \in [0, Q-1],
\end{equation}
where $\nabla w_{t,i}^{0}=\nabla w_{t,i}$. After $Q$-$1$ iterations, the universal adversarial example (perturbed model update), $w_{t,i}^{Q-1}$, can be obtained.
%and $g_q(\cdot)$ denotes the loss function of the substitute model $S_q$. 

\item \noindent\textbf{Meta-test.}
After gaining prior knowledge from the meta-train, the meta-test step is used to fine-tune the perturbation to make it adapt to the unseen model. In this step, we perform the black-box attack against the last sampled model $S_Q$ to improve the generality of the perturbation obtained from the meta-train step. As we cannot access the model's loss function in the black-box setting, we adopt FGSM onto the cross-entropy ($\mathcal{L}_{\textup{CE}}$) between the model predictions and ground-truths to generate perturbations, which can be formulated as:
%Specifically, we estimate the gradient information of the unseen model by using cross-entropy to replace its actual loss function, and the other part works the same. The meta-test can be formulated as follows:
\begin{equation}
\setlength{\abovedisplayskip}{3pt}
\setlength{\belowdisplayskip}{3pt}
    \nabla w_{t,i}^{Q} = \nabla w_{t,i}^{Q-1} + \epsilon \cdot sign(\nabla \mathcal{L}_{\textup{CE}}(S_Q(\nabla w_{t,i}^{Q-1}), a_{i,m})).
\end{equation}
The perturbation generated by the meta-test step ($peb_{test} = \nabla w_{t,i}^{Q} - \nabla w_{t,i}^{Q-1}$) is the defensive perturbation we are seeking for. It relies on the prior knowledge from the meta-train (i.e., using the perturbed model update as the basis) and fine-tuning it to cover the unseen model in the meta-test.
Then we can add it back to the original model update $w_{t,i}$ as follows:
\begin{equation}
\setlength{\abovedisplayskip}{3pt}
\setlength{\belowdisplayskip}{3pt}
    \nabla w_{t,i}^{\prime} = \nabla w_{t,i} + (\nabla w_{t,i}^{Q} - \nabla w_{t,i}^{Q-1})
    = \nabla w_{t,i} + peb_{test}.
\end{equation}

\item \textbf{Avoiding Bias.}
We notice that only performing meta-train and meta-test for one time results in bias to some of the models, and the performance will be highly dependent on the one-time choice. Therefore, we propose to iteratively repeat the two-step process by composing different combinations of various models to improve the transferability further. 
Specifically, \methodName takes the original model update $\nabla w_{t,i}$ as input, and repeats the meta-train/meta-test for $P$ iterations in total. Each iteration takes the output of the last iteration as its input. It can be formulated as follows:
\begin{equation}
\setlength{\abovedisplayskip}{3pt}
\setlength{\belowdisplayskip}{3pt}
    \nabla w_{t,i}^{p\prime} = \nabla w_{t,i}^{p-1\prime} + peb_{test}^{p}, p \in [0,P],
\end{equation}
where $ \nabla w_{t,i}^{0\prime} = \nabla w_{t,i}$ and $\nabla w_{t,i}^{P\prime}$ will be the final perturbation for the single attribute protection, by repeating the meta-train/meta-test steps, the final perturbation will not bias to any model and thus obtain a better transferability to mislead the unseen and unqueyable adversary model.

\end{itemize}
\vspace{-1mm}

%We take the $a_{i,m} \in \mathbf{A}_i$ as the targeted attribute as an example. Before computing perturbation, we need to form a model zoo consisting of multiple pre-trained substitute models $\mathcal{S}$. These models all mimic the behavior of the adversary model, that is, predicting the $a_{i,m}$ through model updates. Then we randomly sample $\{S_1, ..., S_Q\}$ from $\mathcal{S}$ and perform the meta-train and meta-test to compute the generalized adversarial sample. 

%adds perturbation on each iteration and uses the perturbed model update as the input of the next iteration.
%Each iteration includes: (1) randomly samples $Q$ models $\{S_1, ..., S_Q\}$ from the model zoo $\mathcal{S}$, (2) takes the output of last iteration as its input, (3) performs meta-train/meta-test steps to output a perturbed model updates. 

\noindent\textbf{\underline{Multi-Attribute Protection}.}
\label{multiple}
To expand \methodName to protect more than one attribute simultaneously, we will first conduct single attribute protection for each attribute $a_{i,m} \in \mathbf{A}_i$ individually, and then combine all the attribute-specific defensive perturbations by taking into account their protection levels.
%we are able to protect them separately. A simple but useful trick to combine all these perturbations is to weighted average them together as follows:
\begin{equation}
\setlength{\abovedisplayskip}{3pt}
\setlength{\belowdisplayskip}{3pt}
    peb_{t,i} = \sum_{m=1}^{M} \gamma_m \cdot \nabla w_{t,i}^{P,m\prime},
\end{equation}
where $\gamma_m \in (0,1)$ denotes the protection level on the $m$-th attribute given by the client's preference.
Finally, $peb_{t,i}$ is the perturbation to be added on the model update $\nabla w_{t,i}$, which can effectively protect attributes $\mathbf{A}_i$.

\vspace{-2mm}
\section{Experimental Setup}
% In the experiments, we evaluate our defense against two kinds of privacy leakage attacks and compare with five state-of-the-arts defense baselines. The experiments are conducted on a server with two AMD EPYC 7713 CPUs and four Nvidia Quadro A100 GPUs.

%We evaluated attribute inference attack and data reconstruction attack in our work.We evaluate attribute protection under 3 datasets towards 2 settings of adversaries, and compare it with 4 baselines. Also the whole data protection is evaluated under 2 datasets towards 2 kinks of attackers, and compare it with 4 baselines. The experiments are conducted on XXX server.

\subsection{Datasets}
We use four datasets to evaluate \methodName:
(1) \textit{AudioMNIST}~\cite{becker2018interpreting} contains 30,000 audio recordings of spoken English digits, (2) \textit{Adult Income}~\cite{Dheeru2017} contains income records of 48,842 individuals, (3) \textit{Labeled Faces in the Wild (LFW)}~\cite{LFWTech} contains 13,233 facial images from 5,749 people, each image is cropped to $62\times47$ pixels with RGB channels, and (4) \textit{CelebFaces Attributes (CelebA)}~\cite{liu2015deep} contains 202,599 RGB facial images of $32 \times 32$ pixels covering 10,177 identities.

%\begin{enumerate}[leftmargin=*]

%\item  \textbf{AudioMNIST}~\cite{becker2018interpreting} contains 30,000 audio recordings of spoken English digits. 
%\jian{How many attributes in this dataset?}

%\item  \textbf{Adult Income}~\cite{Dheeru2017} contains
%contains $48,842$ samples with $14$ attributes annotations;
%income records of 48,842 individuals.

%\item  \textbf{Labeled Faces in the Wild (LFW)}~\cite{LFWTech} contains 13,233 facial images from 5,749 people, each image is cropped to $62\times47$ pixels with RGB channels.

%\item  \textbf{CelebFaces Attributes (CelebA)}~\cite{liu2015deep} contains 202,599 RGB facial images of $32 \times 32$ pixels covering 10,177 identities.

%\end{enumerate}

Unless stated otherwise, we divide each dataset into $D_{train}$ and $D_{test}$ with $80\%$ and $20\%$ randomly selected samples, respectively.
% We allocate each sample in $D_{train}$ to one client to train the global model collaboratively.
For the FL setting, by default, we assume each client possesses one sample in $D_{train}$. We will study the case where each client possesses multiple data samples in Section~\ref{subsec:converge}. %\luyang{I do not see this covered in Section~\ref{subsec:converge}}.

\vspace{-4mm}
\subsection{FL Model}
% The configurations of the global model for each dataset are as follows:
We adopt different FL model architectures for the four datasets. More details of the architectures can be found in Appendix~\ref{sec:flmodel-arch}.

\noindent (1) \textbf{AudioMNIST}.
We use a CNN model to perform a 10-class classification of the spoken digits (i.e., zero to nine).
We first process the audio into a spectrogram and then apply a model containing three convolutional layers, one hidden layer, and one output layer.
The model is trained on the cross-entropy loss function with a learning rate of $10^{-4}$.

\noindent (2) \textbf{Adult Income}.
We use a 2-layer fully-connected neural network to perform binary classification on the income level (i.e., $>50k$ or not).
The model is trained on the Mean Squared Error (MSE) loss function with a learning rate of 0.01.

\noindent (3) \textbf{LFW}.
We use a CNN model to perform binary emotion classification (i.e., smiling or not).
The model consists of three convolutional layers, one hidden layer, and one output layer.

\noindent (4) \textbf{CelebA}.
We adopt a CNN model to perform binary classification on gender. The model composes three convolutional layers, one hidden layer,  and one output layer and is trained on the cross-entropy loss function with a learning rate of 0.01.

\vspace{-3mm}
\subsection{Defense Settings}
\label{sec:param}
\textbf{\underline{Model Zoo Setting}.}
% For all dataset, the model zoo contains totally $20$ substitute models. All substitute models are a fully-connected neural network with two hidden layers. Because such architecture can achieve satisfactory performance while keep light-weighted to be stored in local devices. 
% We construct the model zoo with $20$ substitute models for all datasets.
% To maintain a minimal model size, all substitute models are configured as 2-layer fully-connected neural networks with varying number of neurons for each layer (i.e., from 128 to 3,000).
To achieve good defense generalizability on various types of adversary models, the model zoo should include a sufficient number of defender models with diverse structures.
Unless mentioned otherwise, in our implementation, we choose to construct the model zoo with $20$ defender models for better transferability and utility-privacy trade-off. We also study the impact of model zoo size in Section~\ref{subsec:computational_cost}.
%\textcolor{red}{for the major results after trying a different number of defender models and found that 20 is the minimum number that can provide desired protection while saving storage and total run time. We also study the impact of model zoo size in Section~\ref{subsec:computational_cost}.}
Our empirical study finds that using defender models with deeper structures (e.g., 4 or 5 layers) would not provide any significant performance benefit over shallow models.
Therefore, to maintain a minimal model size and enable more computational efficiency on clients' local devices, we configure the defender models as 3-layer fully-connected neural networks with varying numbers of neurons for each layer (i.e., 128 to 2,048).

% \zh{diversify, computation cost on client machine, adding more layers do not improve on these tasks}

\noindent \textbf{\underline{Parameter Setting}.}
Unless otherwise stated, we empirically set the number of iterations, $P$, to 10 and the number of selected models for each iteration, $Q$, to 5 for all the datasets to improve the transferability of the defensive perturbations while maintaining a reasonable level of computational cost. %\textcolor{red}{The impact of Q and correspondingly required computational resources are analyzed in Section~\ref{subsec:computational_cost}.}
% We evaluate the perturbation budget $\epsilon \in [5\times 10^{-5}, 0.1]$ for LFW dataset, $\epsilon \in [2\times 10^{-5}, 0.2]$ for Adult Income dataset, and $\epsilon \in [2\times 10^{-5}, 0.2]$ for AudioMNIST dataset.
Additionally, we explore different perturbation budget ranges for different datasets: 
$\epsilon \in [5\times 10^{-5}, 0.5]$ for AudioMNIST dataset;
$\epsilon \in [1\times 10^{-5}, 0.1]$ for Adult Income dataset;
$\epsilon \in [5\times 10^{-5}, 0.5]$ for LFW dataset; and
and $\epsilon \in [5\times 10^{-6}, 0.5]$ for CelebA dataset. We study the impact of privacy budgets in Appendix~\ref{sec:budget}.

\vspace{-4mm}
\subsection{Adversary Models}
\textbf{\underline{Attribute Inference Attack}.}
% The adversary is a classifier trained by the collected gradients and attributes.
The adversary uses a pre-trained classifier to infer sensitive attributes from gradients.
% To reduce the input dimension, the adversary applies the max-pooling operator on the the input gradients following the setting of an existing study~\cite{melis2019exploiting}. We set up the following four kinds of adversary models.
We train the adversary classifiers by the same training set as the defender models (i.e., $D_{train}$) to simulate the strongest adversary who has the knowledge of the defender models' training set.
Following a prior study~\cite{melis2019exploiting}, we assume that the adversary applies max-pooling on the received gradients to reduce dimensionality.
We explore the following adversary model settings:

\noindent (1) \textbf{Structure-known Neural Network (\layerAtt).}
% We assume that the adversary model architecture is included in the defender model zoo. For all datasets, \layerAtt\ uses a fully-connect neural network with two hidden layers.
We assume the adversary model architecture is included in the defender model zoo (i.e., a 3-layer fully-connected neural network).
\layerAtt\ is trained on the cross-entropy loss function using a Stochastic Gradient Descent (SGD) optimizer with a learning rate of 0.01 for 80 epochs.

\noindent (2) \textbf{Unknown Neural Network (\diffAtt).}
We assume that the adversary model architecture is not included in the defender model zoo. It is a 4-layer fully-connected neural network, each layer contains 1,024, 1,024, 512, and 128 neurons, respectively.
% Other settings of \diffAtt\ are the same as \layerAtt.
The training setting of \diffAtt\ is the same as \layerAtt.

\noindent (3) \textbf{Support Vector Machine (\svmAtt).}
% We assume the adversary model is a Support Vector Machine (SVM) classifier. For all dataset, the adversary use the RBF kernel.
We assume the adversary model is a Support Vector Machine classifier with an RBF kernel.

\noindent (4) \textbf{Random Forest (\rfAtt).}
% We assume the adversary model is a Random Forest (RF) classifier. For all dataset, the forest contains 100 trees.
We assume the adversary model is a Random Forest (RF) classifier containing 120 trees.

\noindent\textbf{\underline{Data Reconstruction Attack}.}
Unlike attribute inference attacks, recent studies~\cite{zhu2019deep, zhao2020idlg, yin2021see} showed that the adversary can even completely reconstruct private data by solving a gradient-matching problem. In this paper, we adopt the more advanced method proposed by Geiping \textit{et al.}~\cite{geiping2020inverting} given its capability in reconstructing high-resolution images and robustness against different random initializations. Specifically, the adversary will run 2,000 iterations to match gradients for the data reconstruction.

\vspace{-2mm}
\subsection{Defense Baselines}
\label{sec:defense baselines}
%\zh{could use a table}
% We compare the privacy-utility trade-off of \methodName with five state-of-the-arts baselines.
% As the transformation functions shown in Section X, each of them has a hyperparameter to control the privacy-utility trade-off.
The following state-of-the-art defense baselines (described in Section~\ref{subsec:problem}) are used as baselines. To better compare their utility-privacy trade-offs, we also adjust their defense parameters (level of gradient transformations) in different ranges.

\noindent (1) \textbf{DP (Gaussian)}~\cite{wei2021gradient}. We set $\mu=0$ for all datasets.
We set $\sigma \in [5 \cdot 10^{-5}, 0.5]$,
$\sigma \in [1 \cdot 10^{-5}, 0.1]$,
$\sigma \in [5 \cdot 10^{-5}, 0.5]$, 
and $\sigma \in [5 \cdot 10^{-6}, 0.5]$
for AudioMNIST, Adult Income, LFW, and CelebA datasets, respectively. 
We set $B \in [20, 22]$,
$B \in [20, 22]$,
$B \in [1 \cdot 10^{-8}, 1 \cdot 10^{-3}]$, 
and $B \in [2 \cdot 10^{-6}, 0.1]$
for AudioMNIST, Adult Income, LFW, and CelebA datasets, respectively. 

\noindent (2) \textbf{DP (Laplace)}~\cite{liu2020adaptive}. We set $\mu=0$ for all datasets.
We set 
$b \in [2 \cdot 10^{-5}, 0.5]$,
$b \in [1 \cdot 10^{-5}, 0.1]$, 
$b \in [2 \cdot 10^{-5}, 0.2]$
and $b \in [2 \cdot 10^{-6}, 0.1]$
for AudioMNIST, Adult Income, LFW, and CelebA datasets, respectively. The settings of $B$ follow DP (Gaussian).

%\noindent (2) \textbf{Gradient Clipping}~\cite{geyer2017differentially}. We set 
%$B \in [20, 22]$,
%$B \in [20, 22]$,
%$B \in [1 \cdot 10^{-8}, 1 \cdot 10^{-3}]$, 
%and $B \in [2 \cdot 10^{-6}, 0.1]$
%for AudioMNIST, Adult Income, LFW, and CelebA dataset, respectively.

\noindent (3) \textbf{Gradient Sparsification}~\cite{lin2017deep}. We set
$p \in [20\%, 90\%]$,
$p \in [15\%, 80\%]$,
$p \in [80\%, 99\%]$, 
and $p \in [10\%, 90\%]$
for AudioMNIST, Adult Income, LFW, and CelebA datasets, respectively.

\noindent (4) \textbf{Soteria}~\cite{sun2021soteria}. We set
$p \in [25\%, 95\%]$,
$p \in [20\%, 90\%]$,
$p \in [80\%, 95\%]$, 
and $p \in [50\%, 90\%]$,
for AudioMNIST, Adult Income, LFW, and CelebA datasets, respectively.

\vspace{-3mm}
\subsection{Evaluation metrics}
%We use the following metrics to quantitatively evaluate the utility-privacy trade-off:
%\begin{enumerate}[leftmargin=*]

% \noindent (1) \textbf{Learning Loss.}
% As the utility measurement, it represents the loss value of the global model that is trained by perturbed model updates and evaluated on $D_2$.
\noindent (1) \textbf{Learning Loss}: the loss value of the global model evaluated on $D_{test}$, used as the utility metric. A lower learning loss means better model performance.

% \noindent (2) \textbf{Attack Success Rate (ASR).}
% As the privacy measurement, it represents the number of correct prediction over the total number of model update received by the adversary.
\noindent (2) \textbf{Attack Success Rate (ASR)}: the ratio of correct predictions over the total number of attribute inferences performed by the adversary, used as the privacy metric. A lower ASR means better protection against the attribute inference attack.

\noindent (3) \textbf{Mean Square Error (MSE)}: the pixel-wise mean-square-error between the reconstructed image and the original image. A higher MSE means better protection against the data reconstruction attack.
%\end{enumerate}

%\zh{add metric for the data reconstruction attack (e.g., MSE)}

%\noindent (3)\textit{MSE/FID/Accuracy.}
%xxxxx

\vspace{-3mm}
\section{Experimental Results}

%%%%%%%% Audio mnist %%%%%%%%
\begin{figure*}
    \centering
    \includegraphics[width=0.9\linewidth]{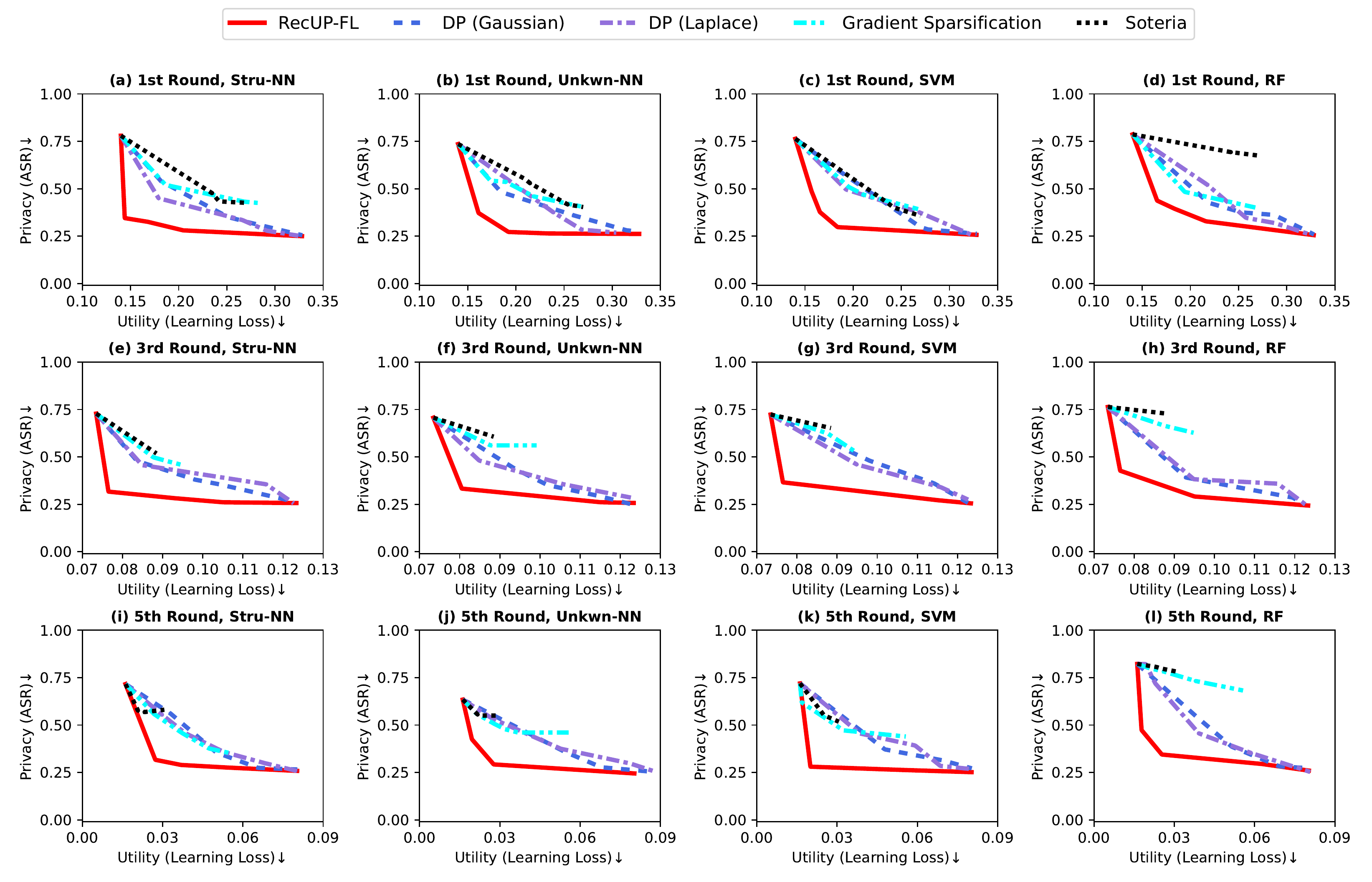}
    \vspace{-4mm}
    \caption{Utility-privacy trade-off curves on AudioMNIST (Some baselines have shorter trade-off curves due to the adjustable range limits of their parameters).}
    % (The length of some curves are limited by their parameters' range.)
    %\zh{Should fix line style and color.} 
    \label{fig:up-audiomnist}
    \vspace{-4mm}
\end{figure*}

%\noindent\textbf{{Settings}.}

\subsection{Single Attribute Protection}
%\jian{Please don't use attribute adversary or data adversary throughout the paper. Please use Attribute Inference Attack and Data Reconstruction Attack.}
%In our experiments, we focus on one communication round each time to obtain a detailed and robust comparison. 

To evaluate \methodName against single-attribute inference attacks, we use three datasets and investigate three typical stages of the FL process, namely, the beginning, the middle, and the end stage of training.
Specifically, according to their convergence speeds, for AudioMNIST, we consider the 1st, 3rd, and 5th rounds with identity (i.e., male native speaker, female native speaker, male non-native speaker and female non-native speaker) as the sensitive attribute; for Adult Income, we consider the 1st, 3rd, and 5th rounds with race (i.e., white, asian-pac-islander, amer-indian-eskimo, black and the other) as the sensitive attribute; and for LFW, we consider the 1st, 10th, and 50th rounds with race (i.e., asian, white and black) as the sensitive attribute.

\noindent \textbf{\underline{Utility-Privacy Trade-off}.}
%\zh{for each defense, we tweak its parameter to get a set of privacy and utility value pairs and plot the privacy-utility trade-off curve accordingly. (1) AudioMNIST: Figure 2 xxxxx; (2)}
%\textcolor{red}{Because we cannot directly compare the privacy budgets based on the parameters in \methodName and other baselines. Instead, }
For each defense, we tweak its parameter as mentioned in Section~\ref{sec:defense baselines} to get a set of privacy and utility value pairs and plot the utility-privacy trade-off curve accordingly. It is worth noting that some curves such as gradient sparsification and Soteria are shorter than others.
% Because they are modifying the gradient itself, the impact of defense is limited within the norm of gradient.
This is because their impacts on the gradients are limited by the parameters' range (e.g., the maximum sparsity is $100\%$).

\noindent (1) \textbf{AudioMNIST}. The results of the AudioMNIST dataset are shown in Figure~\ref{fig:up-audiomnist}.
Since we aim at achieving lower ASR and learning loss, the left bottom corner is the optimal point. Our general observation is \methodName achieves the best utility-privacy trade-off (i.e., closer to the optimal point). For example, when defending against the \layerAtt\ at the 1st round, \methodName can achieve a low ASR ($<30\%$) without scarifying much utility while other baselines keep a relatively large ASR (i.e., around 70\%). When defending against the \diffAtt\ at the 5th round, other baselines can provide sufficient protection only when increasing the perturbation budget to relatively large values, while \methodName can reduce the ASR to around 25\% by adding a much smaller perturbation.

\noindent (2) \textbf{Adult Income}. The results of the Adult Income dataset are shown in Appendix Figure~\ref{fig:up-adult}. We observe that the best trade-off still happens in \methodName. For instance, when defending against the \layerAtt\ at the 3rd round, achieving the same privacy protection (ASR around 35\%), \methodName can keep almost zero utility loss, but other baselines suffer a significant drop of learning loss. However, we notice that when defending against the \rfAtt, our defense performs similarly to the DP (Gaussian) and all baselines cannot get low ASR when adding small perturbation. One of the possible reasons is that \rfAtt~is more robust against the added noise~\cite{ishii2021comparative}.

\noindent (3) \textbf{LFW}. The results of the LFW dataset are shown in Appendix Figure~\ref{fig:up-lfw}. Similarly, we can see that we still achieve the best trade-off.
% For example, when defending against \layerAtt\ at the 1st round, \methodName can perform 12 times better than others on average with the same learning loss (around 0.54), where \methodName decreases the ASR by 73.77\% while others decrease by 5.74\%.
For example, in the case of \layerAtt\ at the 1st round, when achieving the same learning loss (around 0.53), \methodName can largely reduce the ASR by 33.18\% while other baselines can only reduce the ASR by 6.35\%.

In summary, \methodName achieves the best utility-privacy trade-off in the three datasets. This is because \methodName only perturbs the neurons that carry more sensitive information about the specified attributes while preserving the neurons that are relevant to the training task.
% Such perturbation would be less harmful to the FL training while providing better protection against attribute inference attacks.
As a result, the generated perturbation would have less negative effects on the FL training while still being able to provide good protection on the sensitive attributes.

\noindent \textbf{\underline{Transferability of \methodName}.}
Next, we examine the transferability of \methodName by comparing the performance against different adversary models at the same round.

\noindent (1) \textbf{AudioMNIST}.
As shown in Figure~\ref{fig:up-audiomnist}, we can see that \diffAtt~are more difficult to defend than \layerAtt~due to the unknown structure.
% because of the unknown gap.
% But \methodName still achieves lower ASR than other baselines which empirically proves the effectiveness of our method.
\methodName is still able to achieve an ASR that is on average $23.7\%$ lower than other baselines.
Also, we can observe that gradient sparsification can hardly defend against non-neural-network adversaries, especially at the 3rd and 5th rounds. 

\noindent (2) \textbf{Adult Income}.
As shown in Appendix Figure~\ref{fig:up-adult}, \methodName can still achieve the best utility-privacy trade-off regardless of the adversary model architecture. 
%a relatively lower ASR when defending against \diffAtt.
For example, when defending against \diffAtt, \methodName achieves 29.79\% ASR scarifying with almost no learning loss at the 1st round. Also, we can observe that \methodName can effectively defend the SVM adversary on the Adult Income dataset by lowering the ASR to around 28\%, and other baselines can hardly defend them under similar learning loss. 

\noindent (3) \textbf{LFW}.
% As shown in Figure~\ref{fig:up-lfw}, we can also observe the transferability of \methodName by the low ASR in defending against \diffAtt at three stages. Because \rfAtt~has the best robustness against noise, \methodName still slightly outperforms other baselines.  
As shown in Appendix Figure~\ref{fig:up-lfw}, we can observe that \methodName also achieves good transferability on the LFW dataset with the lowest ASR in defending against \diffAtt~at all three stages.
Even when defending against \rfAtt~which has the best robustness against noise, \methodName can still outperform other baselines.

To sum up, these results show that \methodName has the capability in defending against both neural-network and non-neural-network adversary models.
Because the various architectures in the model zoo have a strong ability to fit non-linear mapping functions from gradients to attributes. SVM and RF can also be considered as non-linear functions, and thus the generated perturbation can be utilized to defend against non-neural-network adversary models.

%As shown in Figure~\ref{fig:up-audiomnist}, \methodName can achieve a relatively low ASR compared with baselines in defending SVM and RF adversaries on AudioMNIST dataset.
 %Also, we can observe that gradient clipping and gradient sparsification can hardly defend against the non-neural-network adversaries, especially in the 3-th and 5-th rounds. 
%As shown in Figure~\ref{fig:up-adult}, \methodName can effectively defend the SVM adversary on Adult Income dataset by lowering the ASR to 20\%, and similarly, other baselines can hardly defend them.
%As shown in Figure~\ref{fig:up-lfw}, the trade-off of \methodName is still better than others on LFW dataset.
%From Figure~\ref{fig:up-adult}, gradient clipping and gradient sparsification can slightly decrease the ASR when the perturbation budget gets larger, however, \methodName can 

%The results show \methodName can effectively defend non-neural-network adversaries.

%The results show that the generated perturbation by \methodName can defend non-neural-network adversaries. \methodName can still reach lower ASR than other baselines in all settings even if the attribute adversaries are using SVM and RF classifiers. The results demonstrate the generality of perturbation by \methodName. 
\vspace{-6mm}
\subsection{Multi-Attribute Protection}
%\jian{Need to add justification and some new experiments for protecting more attributes (4-5 attributes) in practice.}
%, which is a practical and common behavior for users.
%\noindent \textbf{Comparison with Baselines.}%The baselines and others work as usual. 

We further evaluate the effectiveness of \methodName on the LFW dataset when the clients specify multiple sensitive attributes.
Specifically, we assume the clients consider their race and age (i.e., baby, child, youth, middle-aged, senior) to be equally important (i.e., $\gamma_{gender} = \gamma_{age} = 0.5$).
% We consider two \layerAtt\  adversaries that aim to predict race and gender to evaluate the ASR separately. 
For the adversary model, we consider two separate \layerAtt\ adversary models that aim to predict race and age to evaluate their ASR respectively.

%%%%%%Multi label baselines%%%%
\begin{figure}
\vspace{-2mm}
    \centering
    \includegraphics[width=0.9\linewidth]{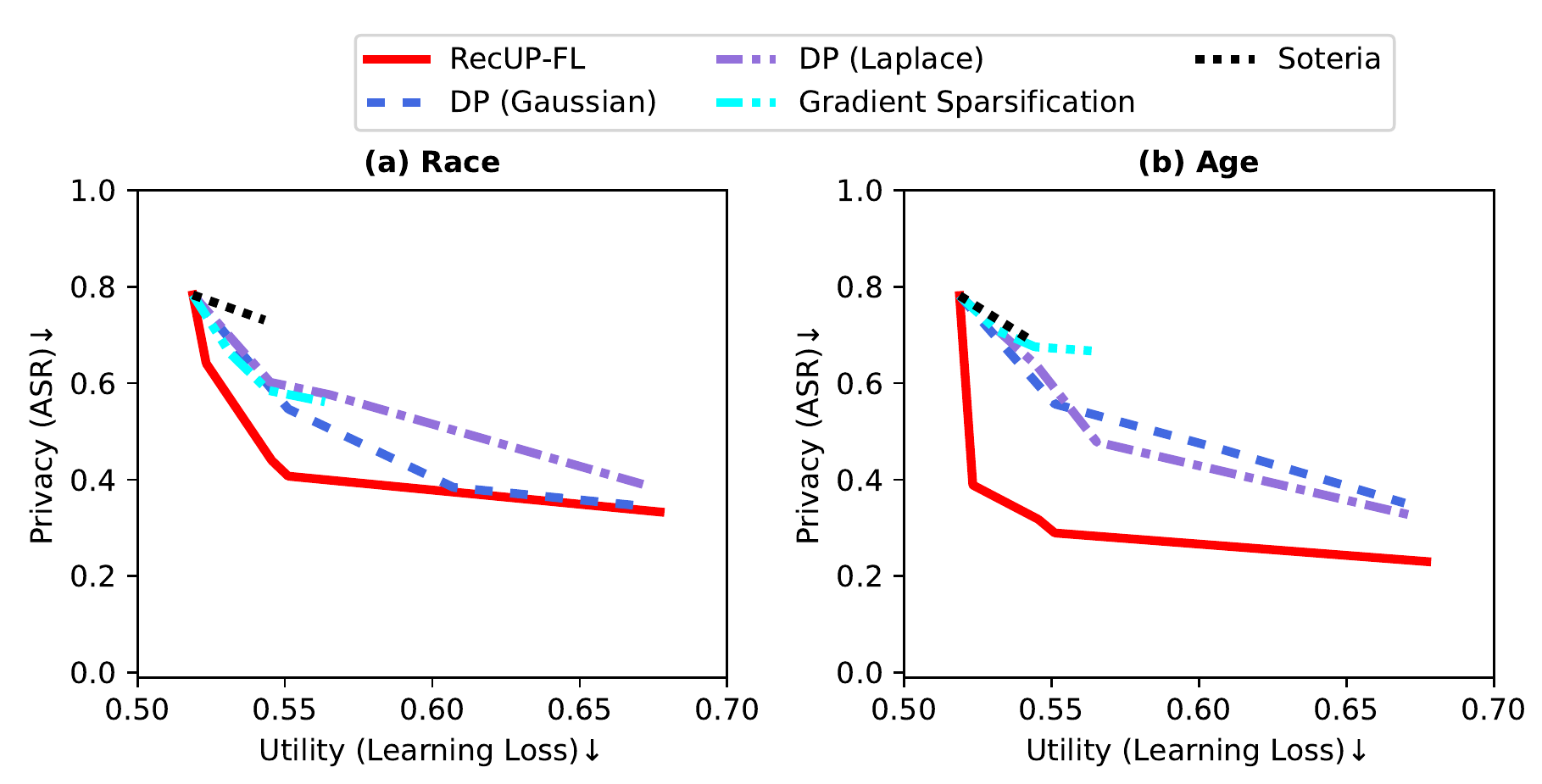}
    \vspace{-4mm}
    \caption{Multi-attribute protection on LFW.}
    \label{fig:lfw-mlabe-lbaselines}
    \vspace{-8mm}
\end{figure}

\noindent \textbf{\underline{Utility-Privacy Trade-off}.}
As shown in Figure~\ref{fig:lfw-mlabe-lbaselines}, compared to other baselines, \methodName can still provide a better level of protection on both attributes under the same utility budget.
For example, in Figure~\ref{fig:lfw-mlabe-lbaselines}(a), we can see that when the learning loss is 0.55, \methodName can achieve an ASR that is 2 times lower than DP (Laplace).
In addition, as shown in Figure~\ref{fig:lfw-mlabe-lbaselines}(b), when the learning loss reaches 0.68, \methodName can achieve around 0.2 ASR while other baselines cannot provide adequate protection on age (ASR > 30\%).

% \zh{add more details}
\begin{figure}
    \centering
    \includegraphics[width=0.94\linewidth]{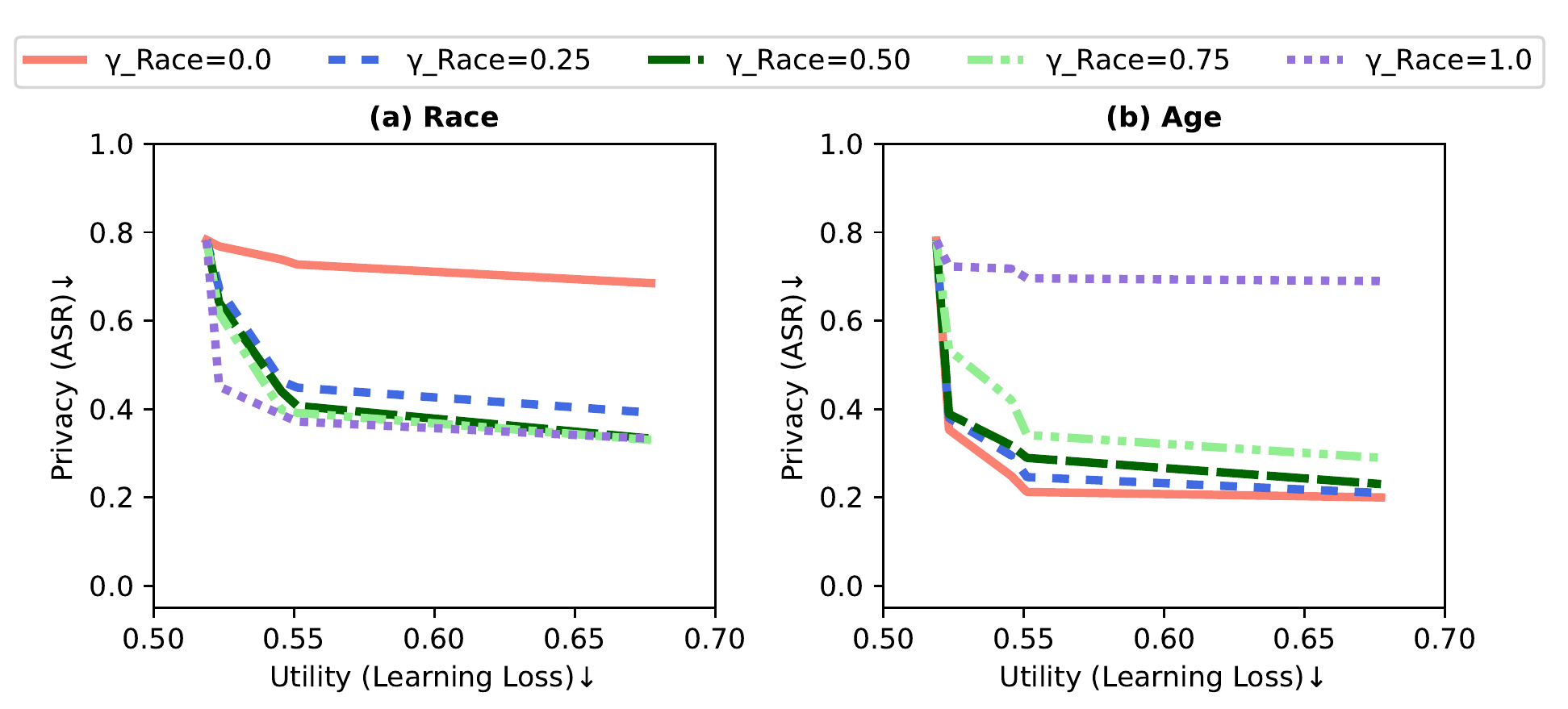}
    \vspace{-4mm}
    \caption{Multi-attribute protection with varying $\gamma$ on LFW.}
    \label{fig:mlabel-ratio}
    \vspace{-6mm}
\end{figure}
\begin{figure}

    \centering
    \includegraphics[width=0.45\linewidth]{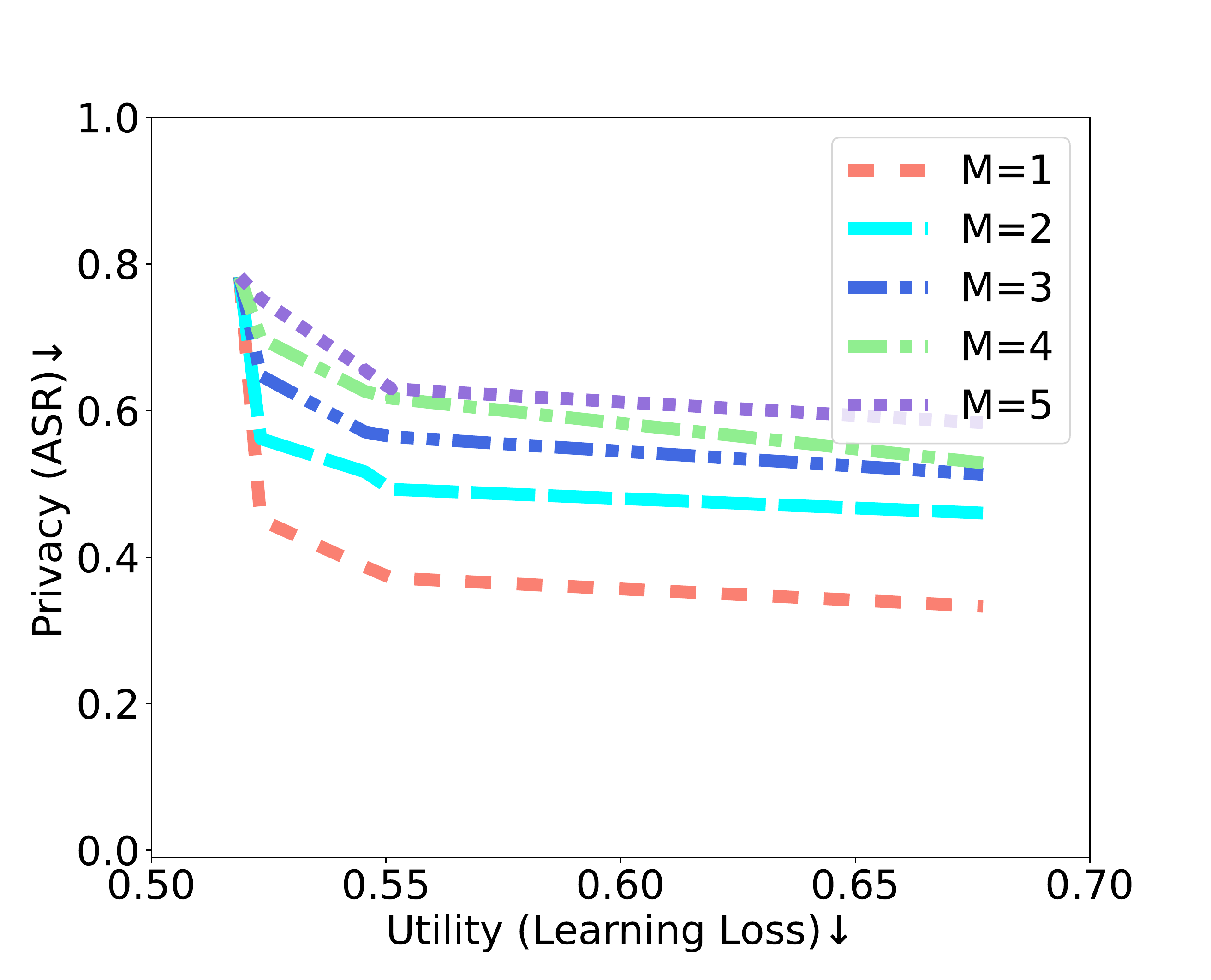}
    \vspace{-5mm}
    \caption{Multi-attribute protection with varying $M$ on LFW.}
    \label{fig:convergence-lfw}
    \vspace{-8mm}
\label{fig:how many attri}
\end{figure}

\noindent \textbf{\underline{Impact of $\gamma$}.}
Following the above-mentioned setting, we set the $\gamma_{race}$ in the range of $[0,1]$ with a step $0.25$ and $\gamma_{age} = 1-\gamma_{race}$.
% As shown in Figure~\ref{fig:mlabel-ratio}, as the $\gamma_{race}$ increases, the ASR of the race attacker drops, and the ASR of the gender attacker increases when achieving the same utility.
As shown in Figure~\ref{fig:mlabel-ratio}(a), given the same utility budget, as we gradually increase $\gamma_{race}$, the ASR against race drops, and the ASR against age increases.
% It is because the increasing $\gamma_{race}$, which means the more allocated perturbation budget on the race, makes the adversary harder to infer its race information.
This is because increasing $\gamma_{race}$ results in more protection over the race attribute while making the age attribute more vulnerable. A similar trend can be observed on the age attribute in Figure~\ref{fig:mlabel-ratio}(b).
%make correct predictions.
% It is worth noticing that even with the $\gamma_{race}=0$, which means that \methodName only aims to protect gender information, the ASR of race still drops slightly as the perturbation budget increases, as shown in Figure~\ref{fig:mlabel-ratio}(a).
We also notice that even with the $\gamma_{race}=0$, the ASR of the race inference attack would still drop slightly when a large perturbation is used.
% Therefore, even \methodName only pays attention to one attribute, it still can provide a certain degree of protection on other attributes because of the correlation of the attributes.
Therefore, even when \methodName is set to focus on a particular attribute, it still has a certain degree of effect over other attributes due to the correlation between these attributes.

\noindent \textbf{\underline{Impact of the Number of Sensitive Attributes}.}
% We explore the maximum number of attributes \methodName can be protected simultaneously. %by gradually increasing the number of attributes $M$ to be protected. 
% \jian{Need more edits. This is not to explore the maximum number of attributes we can protect.}
Next, we conduct experiments on the LFW dataset to study the impact of the number of sensitive attributes $M$ on the performance of \methodName.
% Specifically, we consider the following five sensitive attributes on LFW dataset:
Specifically, we consider emotion (i.e., smiling or not) as the FL training task and investigate the following five sensitive attributes:
(1) \textit{Race}: asian, white and black, (2) \textit{Gender}: male or female, (3) \textit{Age}: baby, child, youth, middle-aged, senior, (4) \textit{Glasses}: eyeglasses, sunglasses, no eyewear, and (5) \textit{Hair}: black hair, blond hair, brown hair, bald.
We assume that the clients treat every attribute equally (i.e., $\gamma = \frac{1}{M}$).
Figure~\ref{fig:how many attri} shows the utility-privacy trade-offs of \methodName against the \layerAtt~adversary that aims to infer the client's race information at the 1st round of FL training.
%\textcolor{red}{More results defending against other adversaries can be found in Appendix Figure~\ref{fig:numattri-lfw}.}
We can observe as we gradually increase the number of protected attributes, the utility-privacy trade-off is becoming worse. 
This is expected since given the same utility budget, protecting a smaller number of attributes allows each attribute to be assigned with a larger weight $\gamma$.
Moreover, when trying to protect a large number (e.g., 5) of sensitive attributes, \methodName still outperforms existing defenses that consider all private information as a single entity compared with the baselines in Appendix Figure~\ref{fig:up-lfw}. For instance, \methodName can decrease the ASR to 0.65 even protecting five attributes simultaneously, while Soteria only reduces the ASR to around 0.75 when they both achieve 0.55 learning loss. These results verify that by selecting a few important sensitive attributes, \methodName can achieve a relaxed form of privacy and thus provide a better utility-privacy trade-off. In some extreme cases, where users want to protect most (or even all) the possible attributes, they can also choose to use traditional privacy defenses to protect the entire data.
% We only consider five attributes because the LFW dataset has only six attribute annotations and we need to leave one of them to be the training task. Also, we believe that choosing five attributes to protect is common and sufficient for users in practice.
% Following the previous setting, we still consider the emotion (i.e., smiling or not) as the FL training task.
% We gradually increase M from 1 to 5 by involving these attributes one by one. For example, M=1 means \methodName only aims at protecting race, M=3 means \methodName is trying to protect race, gender, and age at the same time.
% We evaluate our defense against the \layerAtt~adversary that aims to infer the client's race information at the 1st round of FL training.
% The results are shown in Figure~\ref{fig:how many attri}. We can observe that a smaller M can achieve a better trade-off. This is because, given the same perturbation budget (i.e., achieving the same learning loss), a smaller M means there are fewer attributes required to be protected. And thus the allocated budget on race would become larger, leading to better privacy protection. What's more, we can find that even protecting 5 attributes simultaneously, our defense still performs well. If we combine the left subfigure of Figure~\ref{fig:lfw-mlabe-lbaselines} together to compare, we can find \methodName still outperform other baselines (e.g., \methodName reaches 35.3\% ASR while gaussian noise reaches 67.6\% ASR when learning loss is around 0.7).

%%%%mlabel-ratio
\vspace{-4mm}
\subsection{Defend against Data Reconstruction Attack}
% We investigate three stages of the FL process to evaluate the effectiveness of our defense.
To evaluate the effectiveness of \methodName against data reconstruction attacks, we further conduct experiments on the LFW and CelebA datasets.
Specifically, for the LFW dataset, we consider the 1st, 10th, and 50th rounds with race as the sensitive attribute; and for the CelebA dataset, we consider the 1st, 5th, and 10th rounds with age as the sensitive attribute.
% By tweaking the parameters of each defense, we compare the MSE of each defense when they reach a very close learning loss (within $10^{-3}$).
For a fair comparison, we tweak the parameters of each defense to achieve a similar level of learning loss (within $10^{-3}$).
%To shorten the running time, for each compared defense and communication round, we reconstruct the first 20 images from $D_{test}$ and take the average MSE as the privacy measurement to report.
%We visualize the reconstructed images to compare the performance intuitively.

\begin{figure}
    \centering
    \includegraphics[width=0.95\linewidth]{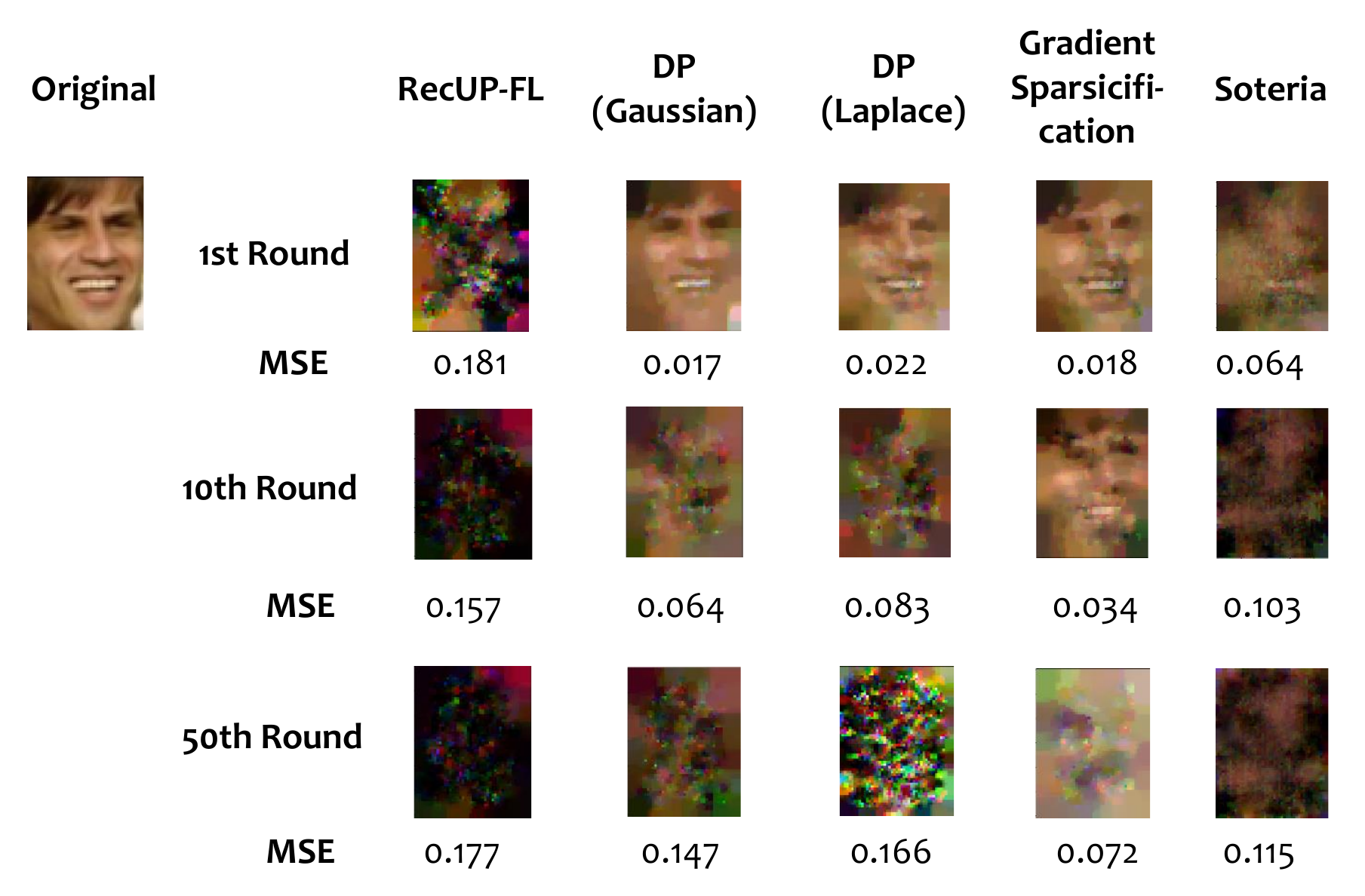}
    \vspace{-4mm}
    \caption{Defending against data reconstruction attack on LFW. }
    \label{fig:data-lfw}
    \vspace{-8mm}
\end{figure}

\noindent (1) \textbf{LFW}. The reconstructed images and the measured MSE on LFW dataset are shown in Figure~\ref{fig:data-lfw}. 
Some visual information can still be revealed from the reconstructions in some situations (e.g., gradient sparsification at the 1st and 10th rounds). However, with \methodName, the reconstructed images do not show any information about specified attributes (i.e., gender).
% , even any facial pattern after applying our defense at all rounds. 
In addition, our defense outperforms other baselines in terms of the measured MSE.
For example, \methodName achieves an 8.2 times greater MSE compared with DP (Gaussian) at the 1st round, a 4.6 times greater MSE compared with gradient sparsification at the 10th round, and a 2.5 times greater MSE compared with gradient sparsification at the 50th round.

\noindent (2) \textbf{CelebA}. The reconstructions and MSE on the CelebA are shown in Appendix Figure~\ref{fig:data-celeba}. 
% Intuitively, we can see the facial pattern after applying Laplacian noise at the 1st and 5th rounds, while no helpful information can be seen after applying our defense at all rounds.
% To quantitatively compare the performance, \methodName achieves 1.8 times larger MSE compared with Laplacian noise at the 1st round, 1.15 times larger MSE compared with Soteria at the 10th round, and 0.7 times larger MSE compared with Soteria at the 50th round.
We observe that the facial structure can still be reconstructed after applying DP (Laplace) at the 1st and 5th rounds.
In comparison, no useful information can be seen from the reconstruction results when \methodName is applied.
Specifically, \methodName can achieve a 3.2 times greater MSE compared with gradient sparsification at the 1st round, a 2.1 times greater MSE compared with DP (Laplace) at the 5th round, and a 2.3 times greater MSE compared with gradient sparsification at the 10th round.

% In summary, the results show that even \methodName is not designed to defend against the data reconstruction attack, \methodName can still provide effective protection against it.
In summary, the results demonstrate \methodName can effectively defend against data reconstruction attacks and achieve better utility-privacy trade-offs compared with other baselines.

\vspace{-2.5mm}
\subsection{Convergence Results}~\label{subsec:converge}
We examine the convergence of the resulting global model when applying \methodName in a non-IID setting with the FedAvg aggregator. We distribute $D_{train}$ of the LFW dataset to 100 clients with no overlap according to their identities, where each client maintains 81 images on average, and evaluate the learning loss of the trained model on $D_{test}$. We consider a practical situation where half of the clients select race as their specified attribute and the other clients select gender. The perturbation budget $\epsilon$ is set to 0.01, while the local training epoch and batch size are set to 1. We evaluate the impact of the participation ratio of clients (ranging from 0.2 to 1.0).
%  Figure~\ref{fig:convergence-lfw} shows that the learning loss decreases as the training goes on.
The convergence results are presented in Figure~\ref{fig:convergence-lfw}. We can see that when \methodName is applied, the model is still able to converge with various participation ratio.
We observe that a lower ratio result in a slower convergence speed as expected because less training data is involved.
%  The lower participation ratio shows a slower convergence speed and larger converged learning loss.
%This is expected because a low ratio involves less training data each round.
In addition, we observe that when the ratio is greater than 0.8, the final learning loss is around 0.42, which is comparable with the case where no defense is applied.
%a no-defense situation.
%  The results show that the global model can converge after applying \methodName even with a low participation ratio (e.g., $ratio=0.2$). 
 %\yue{Should we mention  0.42 loss means  90\% accuracy here? But we haven't mentioned the accuracy before.}
\vspace{-2.5mm}

\subsection{Comparison with Different FGSM Variants}
%\subsection{Effectiveness of Meta-learning}
% We compare our \methodName with four versions of FGSM to evaluate the superiority of leveraging meta-learning.
To verify the effectiveness of the proposed meta-learning-inspired method, we further conduct experiments to compare \methodName with four versions of FGSM under \layerAtt~and \diffAtt.
%Specifically, we compare the following four approaches of FGSM with \methodName 
% The selection of parameters is the same as in Section~\ref{sec:param}. 
We use the same set of parameters as in Section~\ref{sec:param} and the details of each variant can be found in Appendix~\ref{sec:fgsm}. 
As shown in Appendix Figure~\ref{fig:compareFGSM}, all variants can effectively defend against \layerAtt~and \diffAtt, but \methodName achieves the best trade-off among them.
The one-step FGSM achieves the worst trade-offs in all situations as expected. For example, one-step FGSM only achieves around 0.53 ASR while momentum FGSM achieves 0.36 ASR when defending \layerAtt at the 10th round and both of them reach 0.45 learning loss. One possible reason is that perturbation generated by one-step FGSM only relies on one randomly selected model, which is hard to be coincidentally optimal.
%The momentum FGSM achieves even slightly worse trade-offs than one-step FGSM. For example, momentum FGSM only achieves around 20\% ASR, but one-step FGSM achieves around 15\% ASR when defending \layerAtt~at the 10th round, and both of them reach 0.54 learning loss. One possible reason is that the momentum factor limits the extent of perturbation's update each time.
The iterative FGSM and average FGSM perform well, reaching similar ASR when defending \layerAtt~at all three rounds (ASR differences are within 5\%), as they fully utilized all the defender models to generate their perturbation.
%\jian{Add descriptions on the numbers, and move the implementation of those FGSM variants to Appendix.}

\begin{figure}[t]
\begin{minipage}[t]{0.48\linewidth}
    \centering
    \includegraphics[width=0.9\linewidth]{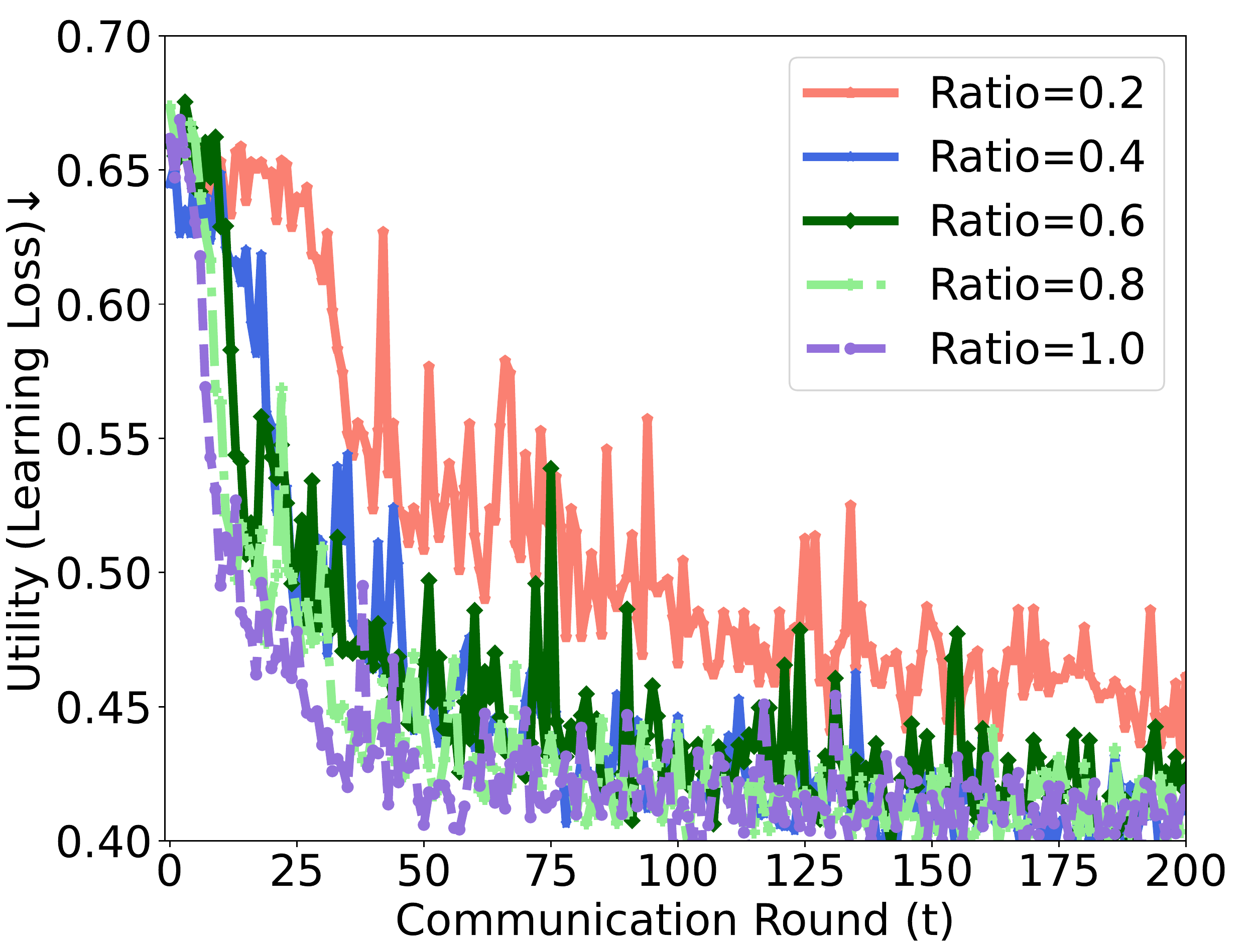}
    \vspace{-4mm}
    \caption{Convergence results on LFW.}
    \vspace{-4mm}
    \label{fig:convergence-lfw}
\end{minipage}
\hspace{0.1cm}
\begin{minipage}[t]{0.49\linewidth} 
    \centering
    \includegraphics[width=0.9\linewidth]{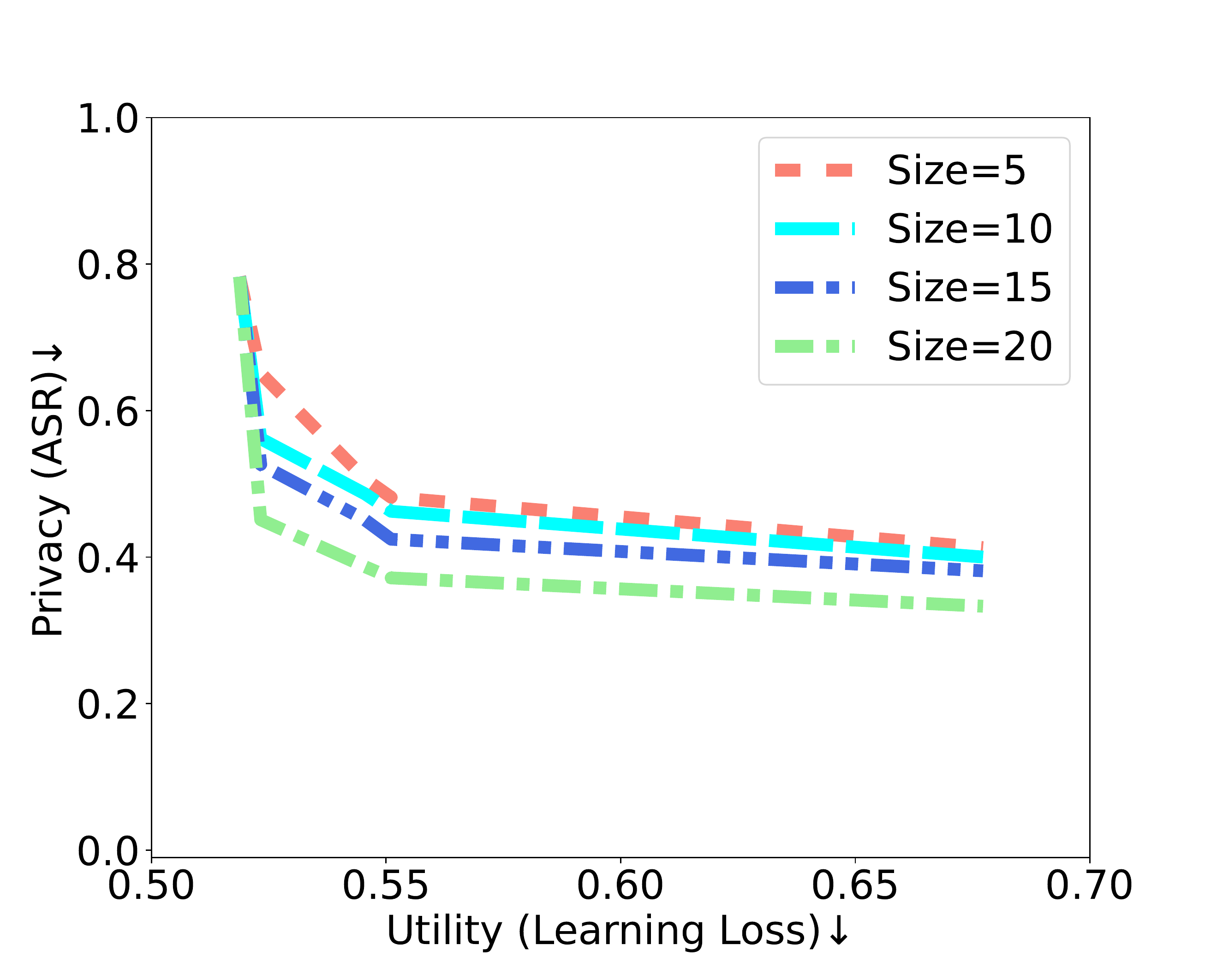}
    \vspace{-4mm}
    \caption{Impact of model zoo size.}
    \vspace{-6mm}
    \label{fig:model zoo size}
\end{minipage}   
\vspace{-1mm}
\end{figure} 

\begin{table}[]
\centering

\caption{Time consumption on defenses.}
\vspace{-4mm}
\renewcommand\arraystretch{1.2}
\resizebox{\linewidth}{!}{
\begin{tabular}{c|ccccc}
\cline{1-6}
\multirow{2}{*}{Defense} & \multirow{2}{*}{\methodName} & \multirow{2}{*}{\begin{tabular}[c]{@{}c@{}}DP\\  (Gaussian)\end{tabular}} &  \multirow{2}{*}{\begin{tabular}[c]{@{}c@{}}DP  \\ (Laplace)\end{tabular}} & \multirow{2}{*}{\begin{tabular}[c]{@{}c@{}}Gradient\\ Sparsification\end{tabular}} &\multirow{2}{*}{Soteria}   \\
&         &                            &   &                    & \\\cline{1-6}
Time (s)    & $7.22\times10^{-2}$   & $5.92\times10^{-3}$                                          & $1.88\times10^{-1}$ & $1.67\times10^{-2}$                                                                                          &  $5.18$                                                                         
\\ \cline{1-6}
\end{tabular}}
\vspace{-5mm}
\label{table:time}
\end{table}

\vspace{-1.5mm}
\subsection{Computational Resource Analysis}
\label{subsec:computational_cost}
\noindent \textbf{\underline{Consumed Time Comparison}.}
We compare average consumed time to apply the defenses per communication round (i.e., \textit{ClientUpdate()} in Algorithm~\ref{algo} line 10 to 25) on LFW dataset using a Nvidia Quadro A100 GPU.
%\jian{I am confused what time are we counting? The perturbation generation time?}\yue{Yes.}
%for each image on LFW after applying defenses.
As shown in Table~\ref{table:time}, our defense achieves a comparable computation time with gradient sparsification and is much faster than Soteria.
%We believe that the latency of \methodName is acceptable (only 72.2 milliseconds) to be deployed on most modern devices.
To evaluate the feasibility of implementing \methodName on users' personal devices such as smartphones and laptops, we estimate the consumed training time via FLOPS (i.e., floating point operations per second). FLOPS measures the computational power of the given hardware and the ratio of FLOPS of two devices can be considered as the ratio of consumed time if they are running the same task~\cite{lu2021fully}. The FLOPS of A100 GPU, popular chips used in smartphones (i.e., Apple A16 Bionic) and laptops (i.e., Intel Core I7 12700H) are $9.7\times10^{6}$ FLOPS~\cite{A100}, $2\times10^6$ FLOPS~\cite{AppleA16} and $1.69\times10^{6}$~\cite{IntelI7} FLOPS, respectively. Thus the estimated consumed time of \methodName on smartphones is 0.35 seconds and 0.41 seconds for laptops, which are still acceptable for edge users who typically have a small-sized local dataset.
Although the training time of some baselines that do not rely on a model to generate the perturbation (e.g., gradient sparsification) is shorter than our defense, they fail to provide a good level of utility-privacy trade-offs.
%adequate protection for the sensitive attributes. 

%In the experiments, we evaluate our defense against two kinds of privacy leakage attacks and compare with five state-of-the-arts defense baselines. The experiments are conducted on a server with two AMD EPYC 7713 CPUs and four Nvidia Quadro A100 GPUs.

%We compare the consumed time for each image in LFW after applying defenses to examine the computational cost. As shown in Table~\ref{table:time}, \methodName can achieve a computation time that is even shorter than the Laplace noise (i.e., $6.03\times10^{-2}$ second faster ), and therefore is sufficient to be run on clients' local devices.
% which is able to be applied to local devices.
%The consumed time of other baselines (except for Soteria) is shorter because they only rely on the random mechanism or the data itself.
%\zh{This statement is false. Laplacian noise is also based on random mechanism. Gradient clipping and sparsification are not random.}
%However, they fail to provide adequate protection for the sensitive attributes. 

\noindent \textbf{\underline{Memory Usage Analysis}.}
% Since our method is designed to be applied on clients' local devices, the memory resource on local devices is limited.
%Memory consumption is also critical for FL defense deployment as it is designed to be applied on clients' local devices.
Memory consumption is also critical for the deployment clients' local devices.
% We investigate the impact of the size of model zoo $\mathcal{S}$ at the AudioMNIST dataset when defending against \diffAtt.
We further conduct experiments on the LFW dataset to investigate the impact of the model zoo size when defending against \diffAtt.
Specifically, we vary the model zoo size from 5 to 20 with a step size of 5. When the size is 5, each iteration shares the same substitute models.
As shown in Figure~\ref{fig:model zoo size}, we can observe that as the model zoo size increases, the trade-off is also improved due to the increased diversity in the model zoo. What's more, even with only five defender models in the zoo, \methodName still outperforms others compared with Figure~\ref{fig:up-lfw}(b). Specifically, when achieving the same learning loss (around 0.55), \methodName with only five defender models reduces the ASR to less than 0.5, while the ASR of DP (Gaussian) and DP (Laplace) is still higher than 0.6.
% However, even when the model zoo only contains 5 substitute models, it only performs 9.5\% worse than containing 20 models on average, proving \methodName is still effective.
%textcolor{red}{We can therefore conclude that even with a small number of models in the model zoo, e.g., five models, \methodName still achieves a relatively good utility-privacy trade-off.}
% For the AudioMNIST dataset, each substitute model is 45 MB on average if they are implemented by Pytorch. Then even if the model size is 20, the total memory size will be 900MB, which is acceptable on most users' devices.
In our Pytorch implementation, each model is \textasciitilde 45MB. Thus \methodName requires \textasciitilde 900 MB even with a large model zoo size of 20, which is still acceptable on most modern devices. The required storage can be further reduced by using a smaller number of models.

%\jian{Need to describe some numbers in text. Can we also describe the required memory size for the model zoom S? Test different number of substitute models in the model zoo.}

%\section{Discussion}
%\jian{We can just discuss membership inference attack.}

%Other than attribute inference attack and data reconstruction attack, membership inference attack is also a popular privacy leakage attack in FL, which determines whether a specific data sample exists in the model's training set or not.
%This attack still leverages a model that mapping a target data and observed gradients to a prediction task~\cite{nasr2019comprehensive, hu2021source} to infer the existence.
%The model is also vulnerable to adversarial attacks, therefore, we can similarly utilize a set of substitute models and iteratively conduct white-box and black-box attack to the adversary.

\vspace{-4mm}

\section{Conclusion and Limitations}
\vspace{-1mm}
\noindent \textbf{\underline{Conclusion}.}
We proposed \methodName, the first user-configurable local privacy defense framework seeking to reconcile the utility and privacy in FL. 
%Unlike existing studies that treat the entire training data as a single entity, we aim to provide effective protection only on user-specific attributes. 
By relaxing the notion of privacy, we focus on the user-specified attributes and thus obtain a significant improvement in model utility.
Inspired by meta-learning, \methodName finds a minimal defensive perturbation to add on the model update before sharing by iteratively conducting white-box and black-box attacks against a set of substitute adversary models. Extensive experiments on four datasets under both attribute inference attacks and data reconstruction attacks show that \methodName can effectively meet user-specified privacy constraints while improving the model utility compared with four state-of-the-art privacy defenses.

\noindent \textbf{\underline{Limitations and Future Work}.}
Our system has the following limitations: 1) Additional Storage Space: As we rely on a set of defender models to generate defensive perturbations, additional storage space ($\sim$hundreds MB) is required on each device. If the device's storage space is not sufficient for the model zoo, we can leverage model compression/quantization techniques to reduce the size of each model; 2) Extra Computational Cost: As the computation of defensive perturbations relies on the proposed two-stage method, extra consumed time ($\sim$0.3 seconds on smartphones) is required. We can leverage the approximate computation to further accelerate \methodName. For example, it is not necessary to get the accurate values of gradients in lines 17 and 20 of Algorithm~\ref{algo} since only the sign of gradient values will be used; 3) Degraded Utility-Privacy Trade-off with a Large Number of Attributes: In some extreme cases, if users want to protect most (or even all) the possible attributes, we can directly apply traditional approaches instead. 
Our future research includes: 1) theoretically deriving certified robustness guarantee and convergence guarantee to FL; and 2) improving the fairness of FL model by reducing its dependency on 
unrelated attributes.

\noindent\textbf{Acknowledgments}
This work is supported in part by NSF CNS-2114161, ECCS-2132106, CBET-2130643, the Science Alliance’s StART program, and the GCP credits by Google Cloud.
%\end{acks}

\vspace{-6mm}
\bibliographystyle{ACM-Reference-Format}
\bibliography{main}

\newpage

\appendix
\section{Introduction of Meta-Learning}
\label{sec:meta-learning}
In traditional machine learning, we decide on a learning algorithm by hand for the desired task and train the model from scratch. However, when the data is expensive or hard to obtain, or computational resources are unavailable, the performance of the traditional scheme will be limited.
%~\cite{ignatov2019ai} 
Meta-learning targets to replace prior hand-designed learners with learned learning algorithms~\cite{hospedales2021meta}. Such a scheme is also called '\textit{learning to learn}'.
%~\cite{thrun1998learning}.
Many definitions and perspectives on meta-learning can be founded in the existing literature. The goal of meta-learning is to learn a model initialization such that it can be quickly adapted to new tasks using limited training samples. Inspired by the intuition of meta-learning, which utilizes prior knowledge learned from a wide distribution of models and adopts it to new tasks, we generate perturbations for the unseen and uncurable adversary models by the proposed two-step iterative method as described in Section~\ref{sec:methodology}.

%Thrun \textit{et al.}~\cite{thrun1998learning} defines it as occurring when the performance at solving tasks drawn from a given task family improves with respect to the number of tasks seen. Finn \textit{et al.}~\cite{finn2017model} defines the goal of meta-learning as training a model on a variety of learning tasks such that it can solve new learning tasks using only a small number of training samples. Nichol \textit{et al.}~\cite{nichol2018first} consider it as where there is a distribution of tasks, we obtain an agent that performs well (i.e., adopt quickly) when presented with a previously unseen task sampled from this distribution. 
%The training process includes three stages: (1) covering a wide distribution of related tasks, (2) distilling the experience, and (3) using the obtained experience to improve future learning performance.
%They propose Reptile, which repeatedly samples a task from the distribution, training on it and moving the initialization towards the trained weights on that task.

%Meta learning has been widely applied to xx such as image recognition~\cite{}, learning fast reinforcement learning agents~\cite{}.

\section{FL Model Architectures}
\label{sec:flmodel-arch}
The detailed FL model architectures of four datasets are shown in Appendix Table~\ref{tab:all}.

\begin{table}[H]
\centering
\caption{FL Models Architecture for four datasets.}
\label{tab:all}

\begin{subtable}[h]{0.45\linewidth}
\centering
\resizebox{\textwidth}{!}{
\begin{tabular}{|c|c|}
\cline{1-2}
\textbf{Layer Type}     & \textbf{Parameters}    \\\cline{1-2}
Input           & $224\times224$                    \\\cline{1-2}
Convolution     & $16\times3\times3$, strides=(2,2)\\
BatchNorm.      & 16                               \\
Activation      & ReLU                             \\
Pooling         & MaxPooling($2\times2$)           \\\cline{1-2}

Convolution     & $32\times3\times3$, strides=(2,2) \\
BatchNorm.      & 32                               \\
Activation      & ReLU                            \\
Pooling         & MaxPooling($2\times2$)          \\\cline{1-2}

Convolution     & $64\times3\times3$, strides=(2,2)\\
BatchNorm.      & 32                               \\
Activation      & ReLU                         \\
Pooling         & MaxPooling($2\times2$)           \\\cline{1-2}
Flatten         &                               \\\cline{1-2}
Fully Connected & 32                              \\
Activation      & ReLU                              \\\cline{1-2}
Fully Connected & 10                              \\\cline{1-2}
\end{tabular}}
\caption{\footnotesize AudioMNIST}
\end{subtable}
\hfill
\begin{subtable}[h]{0.45\linewidth}
\centering
\resizebox{0.9\textwidth}{!}{
\begin{tabular}{|c|c|}
\cline{1-2}
\textbf{Layer Type}     & \textbf{Parameters}    \\\cline{1-2}
Input           & $1\times103$                    \\\cline{1-2}
Fully Connected & 50                              \\
Activation      & ReLU                              \\\cline{1-2}
Fully Connected & 1                              \\
Activation      & Sigmoid                              \\\cline{1-2}
\end{tabular}}
\caption{\footnotesize Adult Income}
\end{subtable}

\begin{subtable}[h]{0.45\linewidth}
\centering
\resizebox{\textwidth}{!}{
\begin{tabular}{|c|c|}
\cline{1-2}
\textbf{Layer Type}     & \textbf{Parameters}    \\\cline{1-2}
Input           & $62\times47$                    \\\cline{1-2}
Convolution     & $32\times3\times3$, strides=(1,1)\\
Pooling         & MaxPooling($2\times2$)           \\
Activation      & ReLU                             \\\cline{1-2}
Convolution     & $64\times3\times3$, strides=(1,1) \\
Pooling         & MaxPooling($2\times2$)          \\
Activation      & ReLU                            \\\cline{1-2}
Convolution     & $128\times3\times3$, strides=(1,1)\\
Pooling         & MaxPooling($2\times2$)           \\
Activation      & ReLU                         \\\cline{1-2}
Flatten         &                               \\\cline{1-2}
Fully Connected & 256                              \\
Activation      & ReLU                              \\\cline{1-2}
Fully Connected & 2                              \\
Activation      & Sigmoid                              \\\cline{1-2}
\end{tabular}}
\caption{\footnotesize LFW}
\end{subtable}
\hspace{\fill}
\begin{subtable}[h]{0.45\linewidth}
\centering
\resizebox{\textwidth}{!}{
\begin{tabular}{|c|c|}
\cline{1-2}
\textbf{Layer Type}     & \textbf{Parameters}    \\\cline{1-2}
Input           & $32\times32$                    \\\cline{1-2}
Convolution     & $16\times3\times3$, strides=(2,2)\\
Pooling         & MaxPooling($2\times2$)           \\
Activation      & ReLU                             \\\cline{1-2}

Convolution     & $32\times3\times3$, strides=(2,2) \\
Pooling         & MaxPooling($2\times2$)          \\
Activation      & ReLU                            \\\cline{1-2}

Convolution     & $64\times3\times3$, strides=(2,2)\\
Pooling         & MaxPooling($2\times2$)           \\
Activation      & ReLU                         \\\cline{1-2}

Convolution     & $128\times3\times3$, strides=(2,2)\\
Pooling         & MaxPooling($2\times2$)           \\
Activation      & ReLU                         \\
Dropout        & 0.2                         \\\cline{1-2}

Flatten         &                              \\\cline{1-2}
Fully Connected & 256                              \\\cline{1-2}
Fully Connected & 128                              \\\cline{1-2}

Fully Connected & 2                              \\
Activation      & Sigmoid                         \\\cline{1-2}

\end{tabular}}
\caption{\footnotesize CelebA}
\end{subtable}

\end{table}

\label{sec:fgsm_results}
\begin{figure}%[H]
    \centering
    \includegraphics[width=\linewidth]{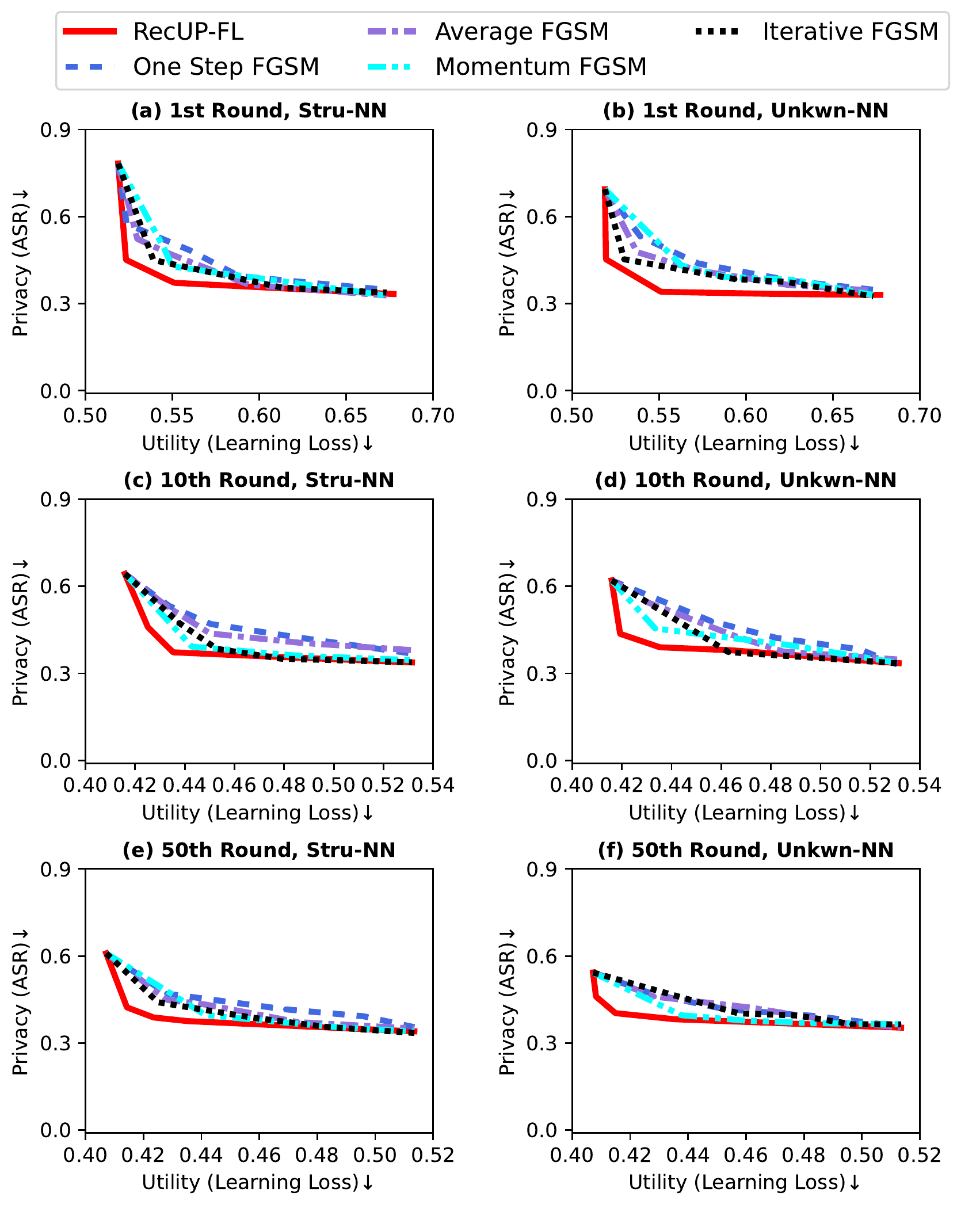}
    \vspace{-8mm}
    \caption{Comparison with four FGSM variants on LFW.}
    \label{fig:compareFGSM}
    %\vspace{-4mm}
\end{figure}

%%%%%%%%Adult %%%%%%%%
\begin{figure*}
    \centering
    \includegraphics[width=0.9\linewidth]{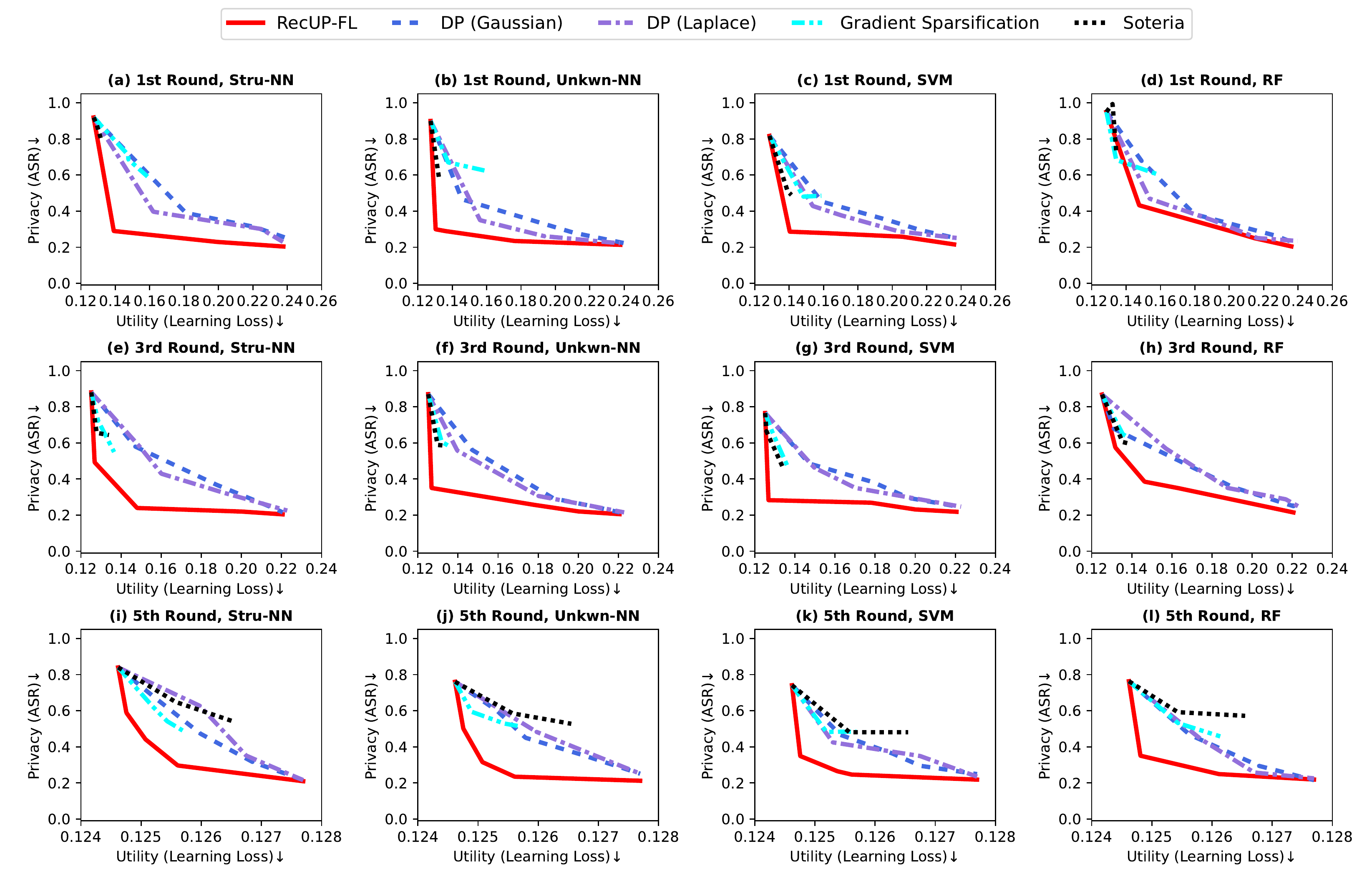}
    \vspace{-4mm}
    \caption{Utility-privacy trade-off curves on Adult Income
    % (The length of some curves are limited by their parameters' range.)
    (Some baselines have shorter trade-off curves due to the adjustable range limits of their parameters).
    }
    \label{fig:up-adult}
    \vspace{-3mm}
\end{figure*}
\section{Utility-privacy Trade-off Curves}
The utility-privacy trade-offs evaluated on the Adult Income and LFW dataset are shown in Appendix Figure~\ref{fig:up-adult} and Figure~\ref{fig:up-lfw}, respectively.

%%%%%%%%LFW %%%%%%%%
\begin{figure*}
    \centering
    \includegraphics[width=0.9\linewidth]{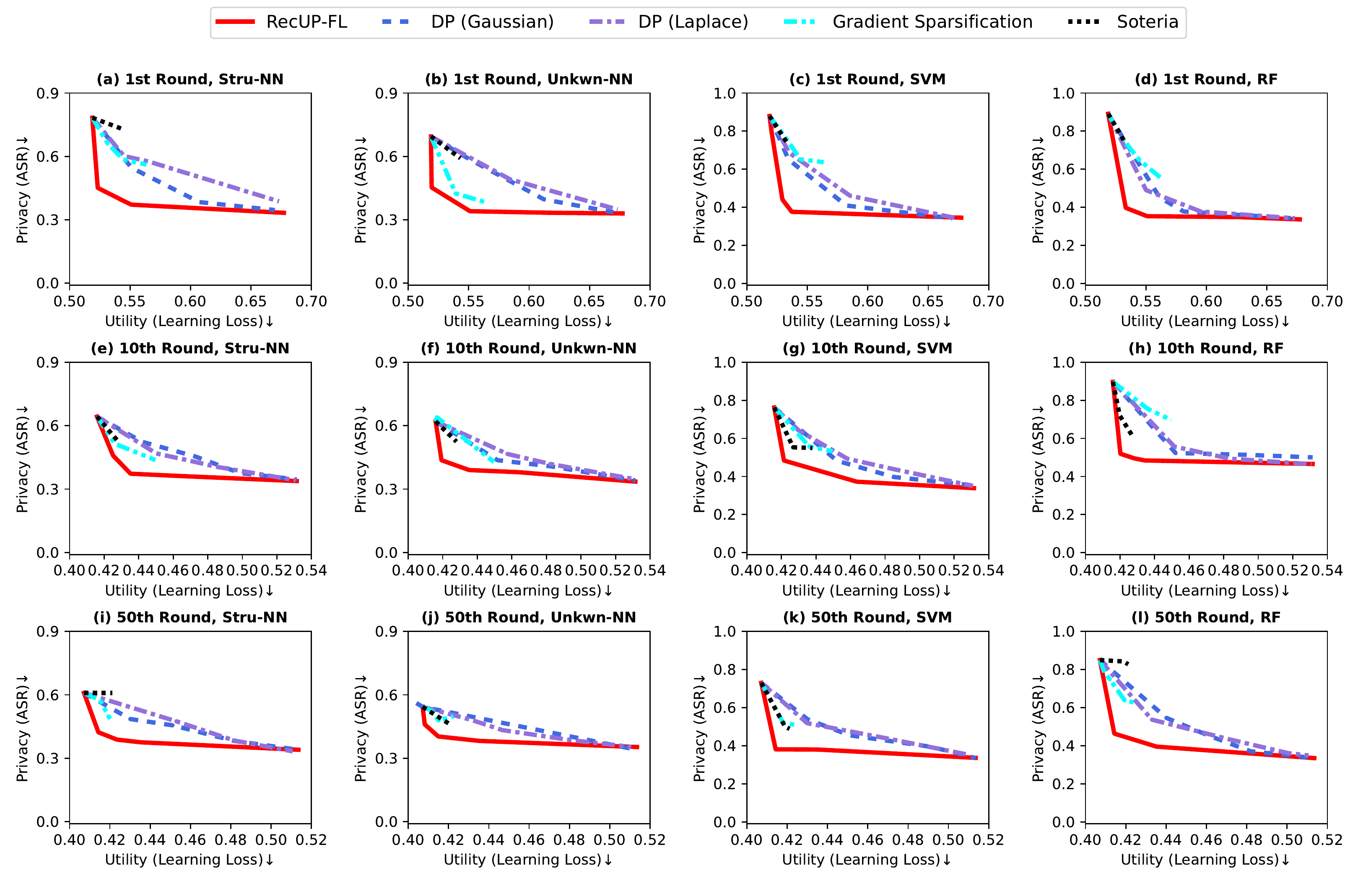}
    \vspace{-5mm}
    \caption{Utility-privacy trade-off curves on LFW
    % (The length of some curves are limited by their parameters' range.)
    (Some baselines have shorter trade-off curves due to the adjustable range limits of their parameters).
    }
    \label{fig:up-lfw}
    \vspace{-8mm}
\end{figure*}

\section{Analysis of Privacy Budgets}
\label{sec:budget}
The utility of global model when applying different privacy budgets of \methodName (i.e., $\epsilon$) at three training stages on LFW dataset are shown in Appendix Figure~\ref{fig:budget-lfw}. We can obviously observe that when the budget increases, the learning loss increases since a larger perturbation is added to the model update.

\begin{figure}
    \centering
    \includegraphics[width=0.55\linewidth]{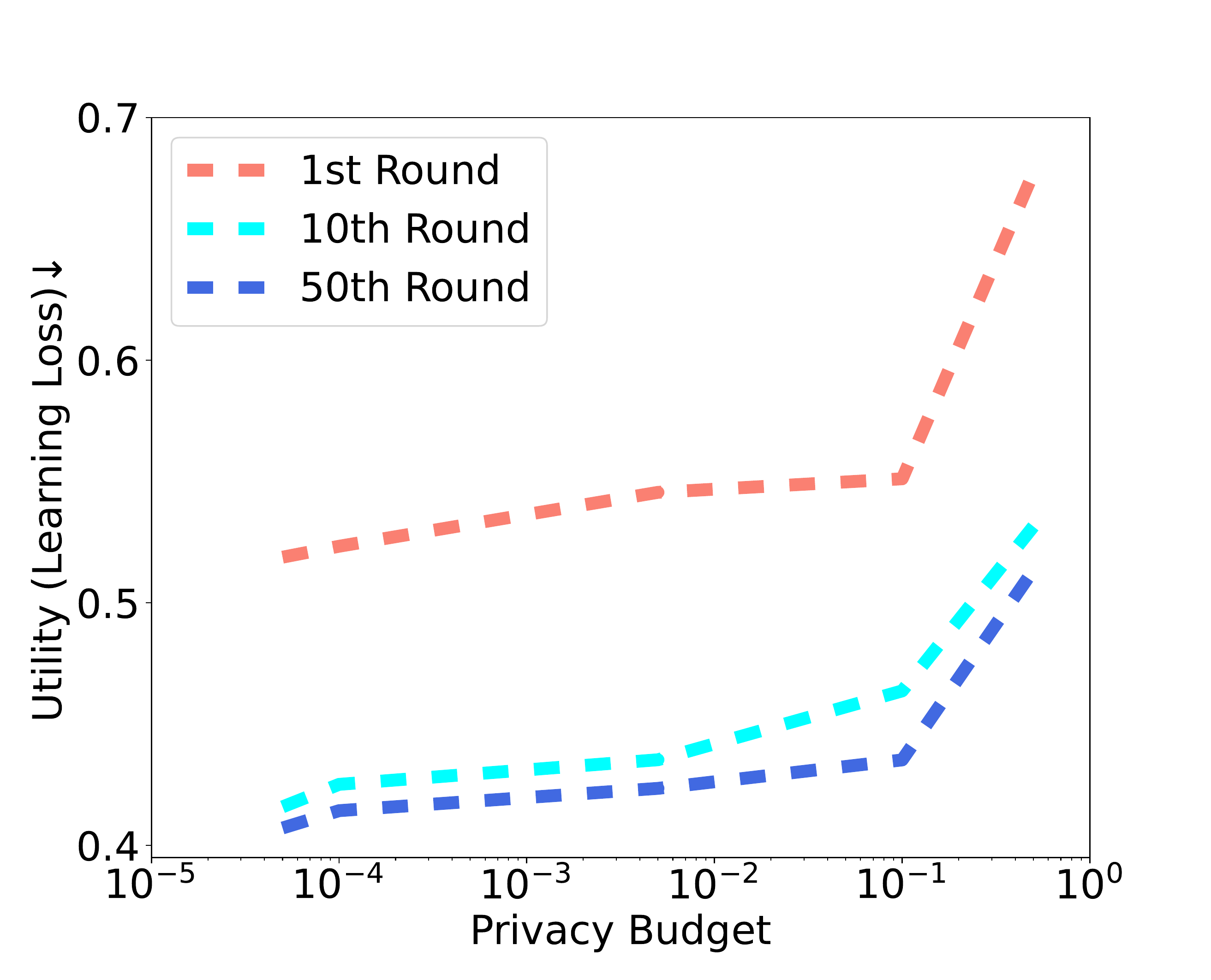}
    \vspace{-4mm}
    \caption{Utilities with varying privacy budgets on LFW. 
    }
    \label{fig:budget-lfw}
    %\vspace{-4mm}
\end{figure}

%\section{Impact of the Number of Sensitive Attributes}
%\textcolor{red}{The utility-privacy trade-off curves of \methodName defending against four adversaries at the 1st round are shown in Appendix Figure~\ref{fig:numattri-lfw}.}
%\begin{figure}
%    \centering
%    \includegraphics[width=1\linewidth]{asiaccs_figs/lfw_num_attri_4all.pdf}
%    \vspace{-8mm}
%    \caption{Multi-attribute protection with varying $M$ on LFW.
%    }
%   \label{fig:numattri-lfw}
    %\vspace{-8mm}
%\end{figure}

\section{Implementation of FGSM Variants}
\label{sec:fgsm}
%\zh{Add a summarizing sentences}
The computation of each FGSM variant is shown as follows:

\noindent (1) \textbf{One Step FGSM.}
We only randomly select one substitute model $S_q$ from model zoo $\mathcal{S}$ and performs FGSM once to get the perturbation as follows:
\begin{equation}
\setlength{\abovedisplayskip}{3pt}
\setlength{\belowdisplayskip}{3pt}
    \nabla w_{t,i}^{\prime} = \nabla w_{t,i} + \epsilon \cdot sign(\nabla g_q(S_q(\nabla w_{t,i}),a_{i,m})).
\end{equation}

\noindent (2) \textbf{Average FGSM.}
We randomly select $Q$ substitute models from model zoo $\mathcal{S}$, performs FGSM separately and averages the perturbations as follows:
\begin{equation}
\setlength{\abovedisplayskip}{3pt}
\setlength{\belowdisplayskip}{3pt}
    \nabla w_{t,i}^{\prime} = \nabla w_{t,i} + \frac{1}{Q} \sum_{q=1}^Q \epsilon \cdot sign(\nabla g_q(S_q(\nabla w_{t,i}),a_{i,m})).
\end{equation}

\noindent (3) \textbf{Iterative FGSM.}
We randomly select $Q$ substitute models from model zoo $\mathcal{S}$, performs FGSM iteratively and accumulates the perturbations as follows:
\begin{equation}
\setlength{\abovedisplayskip}{3pt}
\setlength{\belowdisplayskip}{3pt}
    \nabla w_{t,i}^{q} =  w_{t,i}^{q-1} + \frac{\epsilon}{Q} \cdot sign(\nabla g_q(S_q(\nabla w_{t,i}^{q-1}),a_{i,m})).
\end{equation}

\noindent (4) \textbf{Momentum FGSM.}
We randomly select $Q$ substitute models from model zoo $\mathcal{S}$, performs FGSM iteratively and accumulates the perturbations with a momentum factor $\mu=0.9$ to constrain the direction of the perturbation as follows:
\begin{equation}
\setlength{\abovedisplayskip}{3pt}
\setlength{\belowdisplayskip}{3pt}
    u_{q} = \mu \cdot u_{q-1} + \frac{g_q(S_q(\nabla w_{t,i}^{q-1}),a_{i,m}))}{\|\nabla g_q(S_q(\nabla w_{t,i}^{q-1}),a_{i,m}))\|_1},
\end{equation}

\begin{equation}
\setlength{\abovedisplayskip}{3pt}
\setlength{\belowdisplayskip}{3pt}
    \nabla w_{t,i}^{q} =  w_{t,i}^{q-1} + \frac{\epsilon}{Q} \cdot sign(u_{q}).
\end{equation}
%\newpage

\vspace{-2mm}
\section{Comparison with other FGSM Variants}
The utility-privacy trade-offs of other FGSM variants and our defense on the LFW dataset are shown in Appendix Figure~\ref{fig:compareFGSM}.

\section{Defend against data reconstruction attack}
The reconstructed images and measured MSE values on the CelebA dataset are shown in Appendix Figure~\ref{fig:data-celeba}.

\begin{figure}
    \centering
    \includegraphics[width=0.98\linewidth]{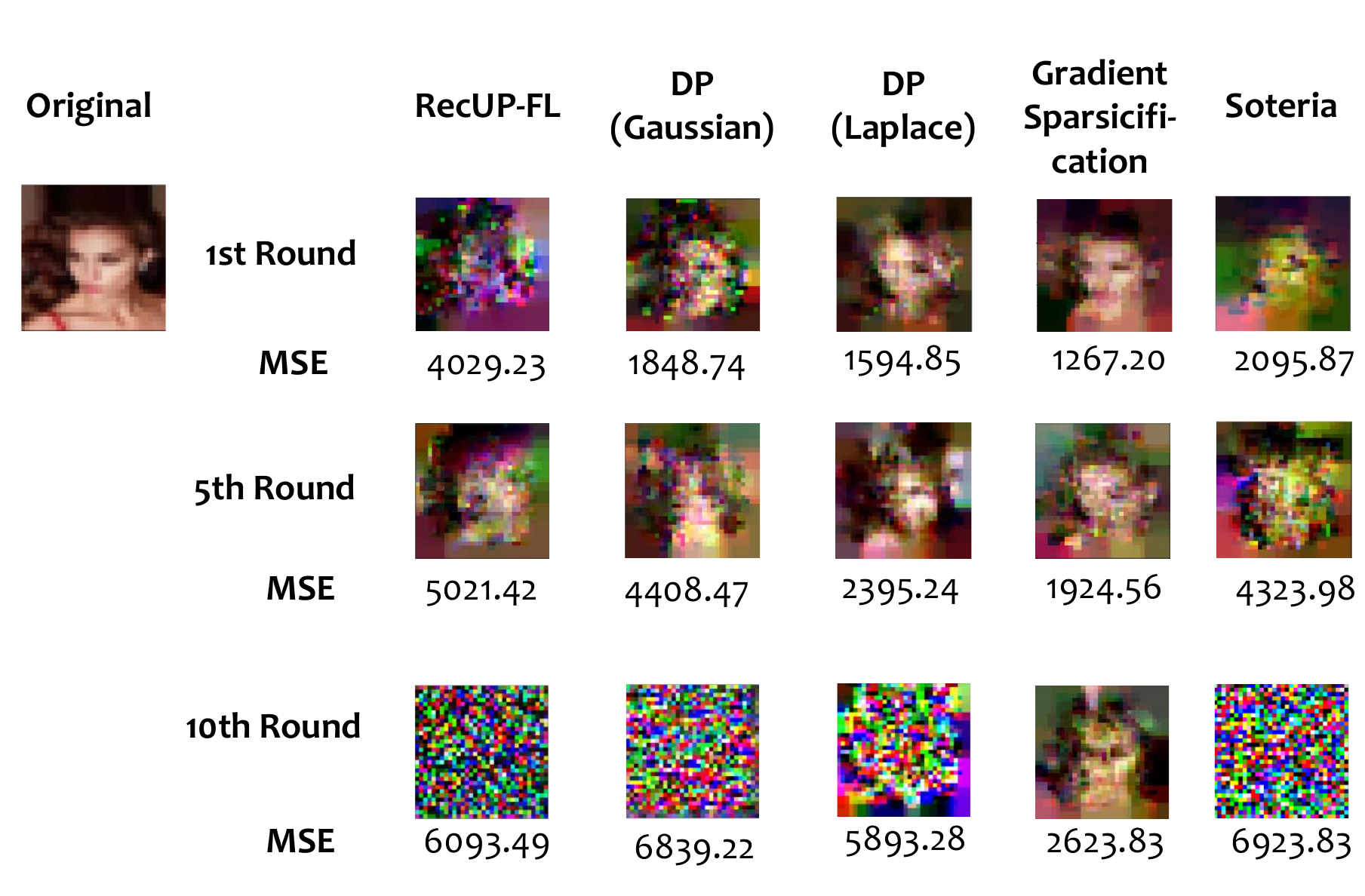}
    \vspace{-2mm}
    \caption{Defending against data reconstruction attack on CelebA.}
    \label{fig:data-celeba}
    \vspace{-4mm}
\end{figure}

\section{Algorithm}
The complete process of generating defensive gradient perturbations can be found in Appendix Algorithm~\ref{algo}.

\begin{algorithm}[t]

\caption{\methodName in FL Pipeline}
\renewcommand{\algorithmicrequire}{\textbf{Input:}}
\renewcommand{\algorithmicensure}{\textbf{Output:}}
 
\begin{algorithmic}[1]
\REQUIRE Learning rate $\alpha$, number of communication rounds $T$
%,  Third-party attacker $A$
\ENSURE Trained global model $\mathcal{G}$ with weight $w_T$

\STATE \underline{\textbf{Server:}}

\STATE Initialize global model $\mathcal{G}$ with weight $w_1$ %\luyang{Either change this to $w_1$ or change the communication round range to [0, T)}
\FOR{each communication round \textit{t $\in$ [1, T]}} 
%\STATE Send global model with weights $w_{t}$ to $K$ clients
%\STATE Select $K$ eligible clients to compute updates
\STATE Randomly select $K$ clients from the entire population
%\luyang{Randomly select $K$ clients from the entire population}

\STATE Send global model with weights $w_{t}$ to $K$ clients
%\luyang{Send global model with weights $w_{t}$ to $K$ clients}

\STATE ${\nabla w_{t,i}}^{\prime} = ClientUpdate(w_{t})$ from the $i$-th client, $i \in [1,K]$ 
\STATE Wait for model updates from $K$ clients %\luyang{I don't think we need this step, or you can move this step below next one}

\STATE Aggregate model updates using FedAvg: $w_{t+1} \leftarrow w_t -\alpha  \frac{1}{K} \sum_{i=1}^{K} {\nabla w_{t,i}}^{\prime}$
\ENDFOR
%\textit{/* Train FL model using original model update*/}
%\OR
%\STATE $w_{t+1} \leftarrow w_t + \alpha \sum g_i^p$ \textit{/* Train FL model using perturbed model update*/}

\STATE \underline{\textbf{ClientUpdate($w_t$):}}
%\STATE The $i$-th client $C_i$ holds its training dataset $\mathbf{D}_i$, configured attributes $\mathbf{A}_i$, and  a perturbation budget $\epsilon$.

%\STATE At round \textit{t=1, 2, ..., T}
%\STATE At the $t$-th round, \textit{t $\in$ [1, T]}:
%\STATE Receive global model $\mathcal{G}$ with weight $w_t$ from the server

%\STATE Execute  $ClientUpdate (w_{t})$ using local data 
%\FOR {$X_{i,n} \in X_i, y_{i,n} \in Y_i$}
\STATE Initialize $peb_{t,i}=0$
\STATE Train model on client's local data $D_i$ and calculate the overall model update by $\nabla w_{t,i} = \frac{\partial \mathcal{L}(D_i, w_t)}{\partial w_t}$ 
%\textit{/* model updates to be protected */} %\luyang{Can we add more context here? Like train model on client's local data and calculate the overall model updates $\nabla w_{t,i}$}

\FOR{each configured attribute $a_{i,m} \in \mathbf{A}_i$}
%\STATE Receive model zoo  $\mathcal{S}$ 
\FOR{iteration \textit{p=1, ..., P}}
    \STATE Randomly select a subset of models $\{S_1, ..., S_Q\}$ from the pre-loaded model zoo $\mathcal{S}$
    %\STATE $input_p = output_{p-1}$
    \STATE \textbf{Meta-train:}
    \FOR{$\textit{q=1,2,...,Q-1}$}
            \item $\nabla w_{t,i}^{q} = \nabla w_{t,i}^{q-1} + \frac{\epsilon}{Q} \cdot sign(\nabla g_q(S_q(\nabla w_{t,i}^{q-1}), a_{i,m}))$
            %\item ${\nabla w_{t,i}} = {\nabla w_{t,i}} + peb_q $
    \ENDFOR
    \STATE \textbf{Meta-test:}
    %\STATE Calculate on the last selected model $S_Q$
    \STATE $\nabla w_{t,i}^{Q} = \nabla w_{t,i}^{Q-1} + \epsilon \cdot sign(\nabla \mathcal{L}_{\textup{CE}}(S_Q(\nabla w_{t,i}^{Q-1}), a_{i,m}))$
    \STATE $\nabla w_{t,i}^{p\prime} = \nabla w_{t,i}^{p-1\prime} + (\nabla w_{t,i}^{Q} - \nabla w_{t,i}^{Q-1})$
    %\STATE $output_p = input_p+peb_Q$
\ENDFOR
\STATE $peb_{t,i} = peb_{t,i} + \gamma_m \cdot (w_{t,i}^{P\prime}-\nabla w_{t,i}) $
\ENDFOR
\STATE return $\nabla w_{t,i}+peb_{t,i}$
\end{algorithmic}
\label{algo}
\end{algorithm}

\section{Theoretical Analysis}
\label{sec:theoretical}
%\noindent\textbf{\underline{Theoretical Analysis}.}
To theoretically show the reason why the perturbation's transferability can be greatly improved through these meta-train and meta-test steps, we consider one iteration and single attribute protection as an example for simplicity. Let $peb_{train}$ denotes the final perturbation generated by the meta-train step, then the objective function of meta-test can be written as:
\begin{equation}
\setlength{\abovedisplayskip}{3pt}
\setlength{\belowdisplayskip}{3pt}
\label{equ: meta-test objective}
    \argmax_{peb_{test}} \mathcal{L}_{\textup{CE}}( S_Q(\nabla w_{t,i}+peb_{train}+peb_{test}), a_{i,m}).
\end{equation}
It means that meta-test tries to find a perturbation for $\nabla  w_{t,i}$ on the basis of the meta-train step to maximize the estimated cross-entropy loss function.
According to the Tayler first-order expansion rule, we can expand the Equation~\ref{equ: meta-test objective} to the following equation:
\begin{equation}
\setlength{\abovedisplayskip}{3pt}
\setlength{\belowdisplayskip}{3pt}
\label{eqa:final}
\begin{aligned}
    \argmax_{peb_{test}}\ &\mathcal{L}_{\textup{CE}}( S_Q(\nabla w_{t,i}+peb_{test}), a_{i,m})+\\
    &peb_{train} \cdot \nabla \mathcal{L}_{\textup{CE}}(S_Q(\nabla w_{t,i}+peb_{test}), a_{i,m}).
\end{aligned}
\end{equation}
To maximize the above objective function, the first term can be considered as the objective function of the meta-test step, which is to mislead the defender model $S_Q$.
The second term can be considered as constraining the gradient directions of $peb_{train}$ and $peb_{test}$ to be as similar as possible.
In other words, the objective function forces meta-test to generate the most similar perturbation as meta-train. It indirectly requires meta-test to utilize the prior knowledge from meta-train and adapt the perturbation to the new task, which is consistent with our design.

\newpage

\end{document}